%% file: 00_main.tex
\pdfoutput=1 
\documentclass[10pt,twocolumn,letterpaper]{article}
\usepackage{cvpr}
\usepackage{times}
\usepackage{epsfig}
\usepackage{graphicx}
\usepackage{amsmath}
\usepackage{amssymb}
\usepackage{url}            
\usepackage{booktabs}       
\usepackage{amsfonts}       
\usepackage{nicefrac}       
\usepackage{microtype}      
\usepackage[noadjust]{cite}
\usepackage{amsmath,amssymb}
\usepackage{textcomp}
\usepackage{gensymb}
\usepackage{algorithm}
\usepackage[algo2e]{algorithm2e}
\usepackage{color}
\usepackage{amsfonts}
\usepackage{graphicx}
\usepackage{tabulary,multirow,overpic,xcolor}
\usepackage{wrapfig}
\usepackage{blindtext}
\usepackage[super]{nth}
\usepackage[acronym]{glossaries}
\usepackage{siunitx}
\usepackage{adjustbox}
\usepackage{makecell}
\input{variables}

\usepackage{subfigure}
\usepackage{graphicx}
\usepackage{tabularx}
\usepackage{subfigure}
\usepackage{fancyhdr} 
\usepackage{bbm}

\usepackage[pagebackref=true,breaklinks=true,colorlinks,bookmarks=false]{hyperref}

\def\cvprPaperID{7655} 

\newacronym{knn}{KNN}{$k$-nearest neighbor}
\newacronym{dnn}{DNN}{deep neural networks}
\newacronym{rnn}{RNN}{recurrent neural networks}
\newacronym{cnn}{CNN}{convolutional neural networks}
\newacronym{vae}{VAE}{variational autoencoder}
\newacronym{rl}{RL}{reinforcement learning}
\newacronym{ssm}{SSM}{state-space model}
\newacronym{air}{AIR}{attend-infer-repeat}
\newacronym{gnn}{GNN}{graph neural networks}
\newacronym{gmm}{GMM}{Gaussian mixture model}
\newacronym{elbo}{ELBO}{evidence lower bound}
\newacronym{gatsbi}{GATSBI}{Generative Agent-cenTric Spatio-temporal oBject Interaction}

\cvprfinalcopy 

\def\cvprPaperID{7655} 

\begin{document}
\title{\LARGE \bf GATSBI: Generative Agent-centric Spatio-temporal Object Interaction}

\author{
Cheol-Hui Min\hspace{1.0cm} Jinseok Bae\hspace{1.0cm} Junho Lee\hspace{1.0cm} Young Min Kim\\
\\
Dept. of Electrical and Computer Engineering, Seoul National University, Korea
\\
{\small \texttt {\{mch5048, capoo95, twjhlee, youngmin.kim\}@snu.ac.kr}}
}
\maketitle
\thispagestyle{empty}
%
\begin{abstract}
We present GATSBI, a generative model that can transform a sequence of raw observations into a structured latent representation that 
fully captures the spatio-temporal context of the agent's actions.
In vision-based decision making scenarios, an agent faces complex high-dimensional observations where multiple entities interact with each other. 
%
The agent requires a good scene representation of the visual observation that discerns essential components and consistently propagates along the time horizon.
%
Our method, GATSBI, utilizes unsupervised object-centric scene representation learning to separate an active agent, static background, and passive objects.  
GATSBI then models the interactions reflecting the causal relationships among decomposed entities
and predicts physically plausible future states.
Our model generalizes to a variety of environments where different types of robots and objects dynamically interact with each other.
We show GATSBI achieves superior performance on scene decomposition and video prediction compared to its state-of-the-art counterparts.
\end{abstract}


\input{01_intro.tex}
\input{02_related_work.tex}
\input{03_gatsbi}
\input{04_experiments}

\input{05_conclusion}
\input{06_acknowledgement}

{\small
\bibliographystyle{ieee_fullname}
\bibliography{00_main}
}

\newcommand{\hiddensubsection}[1]{
    \stepcounter{subsection}
    \subsection*{\arabic{chapter}.\arabic{section}.\arabic{subsection}\hspace{1em}{#1}}
}
\newcommand{\hiddensection}[1]{
    \stepcounter{section}
    \section*{\arabic{chapter}.\arabic{section}.\arabic{subsection}\hspace{1em}{#1}}
}

\newpage
\onecolumn
\renewcommand*\contentsname{Contents (with the sections of the main paper)}
\tableofcontents
\newpage
\input{supplementary}

\end{document}

%% file: variables.tex
\newcommand*{\E}{\ensuremath{\mathop{\mathbb{E}}}}

\newcommand{\kl}[2]{\ensuremath{\mathrm{D_{KL}}\left(#1\;\middle\|\;#2\right)}}

\newcommand{\MLP}{\text{MLP}}
\newcommand{\CNN}{\text{CNN}}
\newcommand{\LSTM}{\text{LSTM}}

%% file: 01_intro.tex
\section{Introduction}
\label{sec:introduction}

An ideal intelligent agent should be able to learn various tasks in diverse environments without relying on specific sensor configurations or control parameters.
Recent approaches employ visual observation as the sole input to infer the physical context of the agent and its surroundings, thus aim to adapt to a general setup.
One may interpret the visual input via conventional computer vision techniques employing deep neural networks~\cite{he2017mask, redmon2018yolov3}.
While they exhibit performance comparable to human perception, such approaches require a large volume of annotated database.
Not only are the groundtruth labels costly to obtain, but also such supervised approaches are limited to specific tasks that they are trained on.

\input{fig/intro/00_intro}

In contrast, unsupervised generative models extract the latent variables that encode the compositional relationship between different entities without prior knowledge~\cite{kingma2013auto}.
The quality of representation can be verified by the ability of reconstructing the input video sequence from the disentangled latent variables~\cite{chung2015recurrent}.
In an ideal case, the latent variables contain the time-varying composition between the agent and the set of objects, and the structural knowledge must propagate temporally with a consistent inference of the latent dynamics. 
The latent dynamics of the learned representation reflect the underlying physics between the extracted entities, thus the agent can leverage the latent dynamics in predicting the various physical contexts conditioned on its own action. 


We propose a fully-unsupervised action-conditioned video prediction model, named \acrfull{gatsbi}.
Our method is explicitly designed for vision-based learning of robot agents and is able to distinguish the active, passive and static components from the robot-object interaction sequence, Fig.~\ref{fig:intro}. 
Conditioning only on actions and a few frames, the learned latent dynamics can predict the long-term future observations without any prior labels of individual components or physics model. 

Our generative model sequentially factorizes each video frame into individual components and extracts the latent dynamics. 
Specifically, our unsupervised network first models relatively large scene components as $2$D-\acrfull{gmm}.
In addition, a group of $2$D-Gaussian keypoints captures actively moving pixels in response to the given action.
One of the GMM modes that matches best with the keypoint-based representation is selected and refined to learn the latent dynamics of the active agent.
In the meantime, small passive objects are extracted by attention-based object discovery models \cite{lin2020improving}.
Finally, \acrfull{gnn}~encodes the interactions between the active agent, passive objects, and static background that are disentangled in the extracted latent variable.
The three different categories of the scene entities are reflected as inherent physical properties within the graphical model, which correctly updates the state of each object in response to the diverse interactions.

In summary, GATSBI is an unsupervised representation learning framework that infers a decomposed latent representation of the observation sequence and predicts associated latent dynamics in an agent-centric manner. 
GATSBI can distinguish various components and correctly understand the causal relationship between them from a sequence of visual observations without specific labels or prior.
Being able to locate the active agent, the acquired latent representation is aware of the dynamics in response to the control action, and can readily be applied to an agent in making physically-plausible decisions. 
We provide extensive investigation on both qualitative and quantitative performance of GATSBI for video prediction on various robot-object interaction scenarios. 
We also compare our model with previous methods on spatio-temporal representation learning and show their promise and limitation.

%% file: fig/intro/00_intro.tex
\begin{figure}[!t]
    \vspace{-0.10cm}
    \centering
    \includegraphics[scale=0.195]{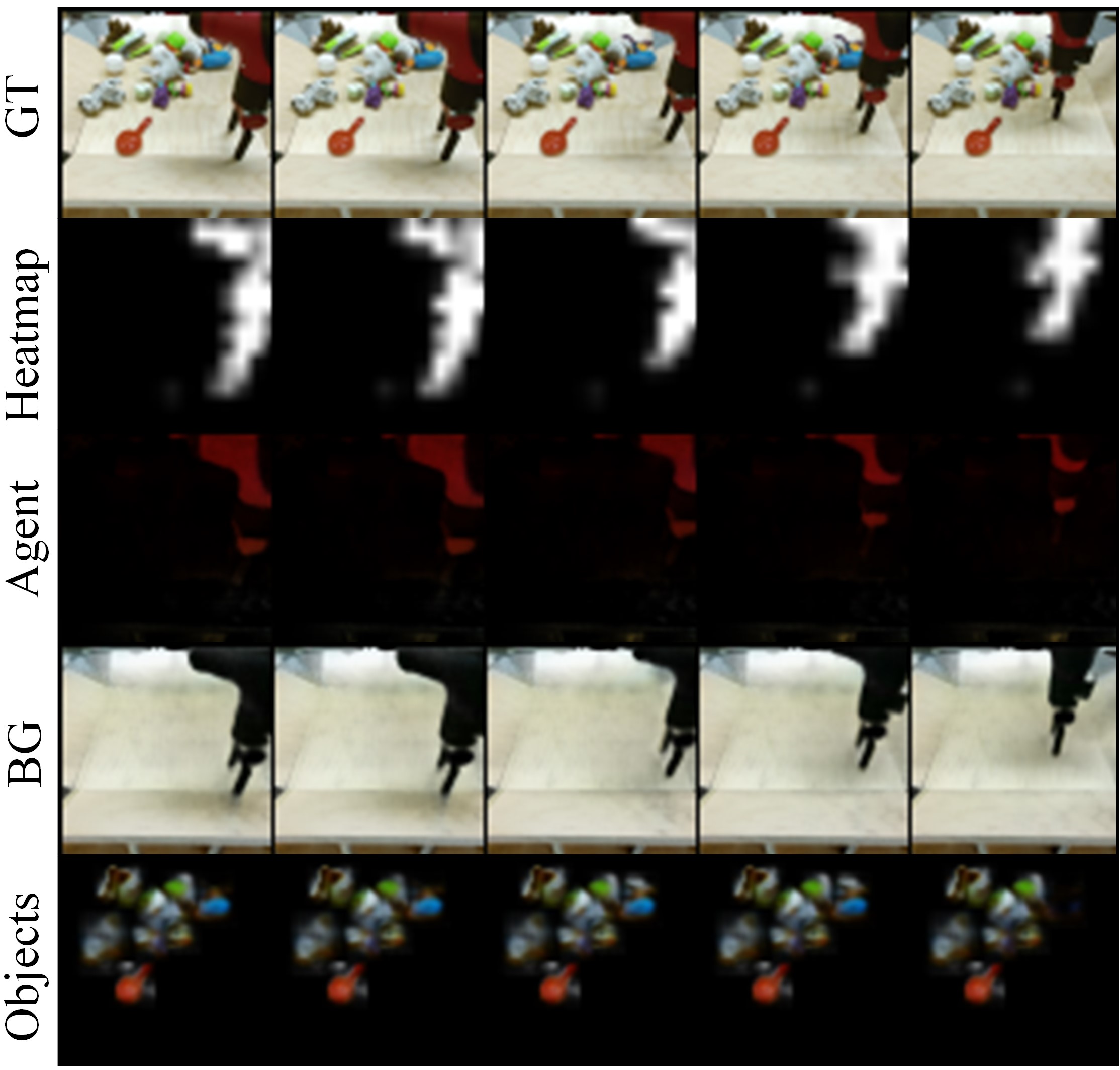}
    \caption{
    Our method, GATSBI, can explicitly identify the agent by utilizing the keypoint-based heatmap. Thus, the observation is decomposed into the agent, background, and objects. In addition, GATSBI infers the dynamic properties of the agent and temporally models the agent-centric interaction with the objects.
    }
    \label{fig:intro}
    \vspace{-0.6cm}
\end{figure}

%% file: 02_related_work.tex
\section{Related Work}
\label{sec:relatedwork}

\subsection{Object-centric Representation Learning}
\label{subsec:ool}

Deep generative models project the high-dimensional visual observation into low-dimensional latent representations~\cite{kingma2013auto,srivastava2015unsupervised,chung2015recurrent}. 
Especially, object-centric representation learning extracts a structured representation that can be mapped into semantic entities.

The representation can be grouped into three categories depending on how the entities are distinguished: attention-based, spatial mixture-based, and keypoint-based methods.
Attention-based methods~\cite{eslami2016attend,kosiorek2018sequential,stelzner2019faster,crawford2019spatially,crawford2019exploiting,stanic2019r} use spatial attention for object discovery and capture the locally consistent spatial features. 
They are good at detecting a large number of scene entities that are confined to small segments of the scene~\cite{crawford2019spatially}.
Spatial mixture methods~\cite{greff2017neural, burgess2019monet, greff2019multi, engelcke2019genesis}, on the other hand, represent relatively large scene entities with Gaussian-mixture model (GMM).
In contrast to the attention-based method, they struggle with scaling to a larger number of scene entities \cite{lin2020space}.
Keypoint-based methods~\cite{minderer2019unsupervised,kulkarni2019unsupervised} extract keypoints from feature maps in an unsupervised way and are recently getting attention for their flexibility in representation.
All of the three approaches have different capabilities of representation, and we carefully coordinate them to correctly disentangle various scene components in an unsupervised setting.
As a concurrent work, \cite{bear2020learning} augments typical \acrfull{cnn} with a graph architecture to find the scene structure during the pixel encoding process.
\subsection{Latent Dynamics Model from Visual Sequence}
\label{subsec:latent_dyna}

The latent representation with discovered objects can be extended to model the temporal transition and interaction of the detected objects~\cite{kosiorek2018sequential, van2018relational, stelzner2019faster,crawford2019spatially, crawford2019exploiting, stanic2019r,jiang2020scalor,kossen2019structured}.
Specifically, \acrfull{ssm}~\cite{chung2015recurrent, karl2016deep, fraccaro2016sequential, doerr2018probabilistic} utilizes \acrfull{rnn}~\cite{hochreiter1997long,chung2014empirical} to pass latent information over a long time sequence, then a graph neural network (GNN) is used to model the interaction between entities \cite{kossen2019structured}.
Concurrent works temporally extend spatial mixture model to achieve the same objective~\cite{zablotskaia2020unsupervised, weis2020unmasking}.
The aforementioned works use object-centric representation to model passive dynamics within the scene, but do not model the intelligent agents.

On the other hand, several recent works incorporate the action (i.e., control command) of the agents in the latent representation~\cite{watter2015embed, finn2016deep, babaeizadeh2017stochastic, ha2018world, hafner2018learning, hafner2019dream,  lee2019stochastic, zhang2019solar, veerapaneni2019entity}.
%
%
While one can use the convolutional recurrent neural network to embed the entire past observations and actions to yield rich temporal information \cite{babaeizadeh2017stochastic},
most works first extract the low-dimensional latent dynamics model from the observation with action-conditioned SSM, and integrate the learned latent model into the agent's policy~\cite{lee2019stochastic} or vision-based planning~\cite{hafner2018learning}.
However, these approaches use a simple variational autoencoder to extract the latent state and thus cannot represent entity-wise interaction.
Previous approaches using structured representation in control tasks either do not detect active agents~\cite{veerapaneni2019entity} or are not tested on the scenes with agents~\cite{watters2019cobra,lin2020improving}.
Compared to these approaches, GATSBI learns representation that explicitly locates active agents and is suitable for learning the different physical properties of agent-object interactions or that of object-object interactions. 

%% file: 03_gatsbi.tex
\section{GATSBI: Generative Agent-centric Spatio-temporal Object Interaction}
\label{sec:gatsbi}
\input{fig/schemes/00_background} 

Given a sequence of observation $o_{0:T}$ and action $a_{0:T}$, GATSBI is designed to embed individual frames into a set of decomposed latent variables $z_t$ which allows us to explicitly represent the dynamics of the agent and resulting entity-wise interactions within the latent space.

Our representation of the observation bases on the variational autoencoder (VAE)~\cite{kingma2013auto} that encodes the high-dimensional visual observation $o$ into a low-dimensional latent variable $z$ sampled from a probabilistic distribution. 
We can approximate the probability distribution of the observation $p_\theta(o)$
by maximizing the following empirical lower bound,
\begin{equation}
        \label{eq:elbo_1}
        \log p_\theta(o)  \geq 
         \E\left[\log p_\theta(o|z) \right] - \kl{q_\phi(z|o)}{p_\theta(z)}.
\end{equation}
The lower bound on the right side of the inequality is the evidence lower bound (ELBO)~\cite{kingma2013auto} and optimized with neural networks parameterized by $\theta$ and $\phi$.
$p_\theta(o|z)$, $q_\phi(z|o)$, and $p_\theta(z)$ represent the observation likelihood, posterior distribution, and the prior distribution, respectively.

Adopting the state-space model (SSM)~\cite{karl2016deep}, we can temporally expand the basic VAE as in Fig.~\ref{fig:rssm}.
Given a sequence of observation $o_{0:T}$ and the action $a_{0:T}$, the set of structured latent embedding $z_{0:T}$ is also defined as a temporal variable.
In order to maintain the consistent structured representation of the complex observation sequence, RNN memorizes the information to the hidden state $h_{0:T}$,
\begin{equation}
h_t=\mathrm{LSTM}(z_{t-1}, \mathrm{CNN}(o_{t-1}), h_{t-1}).
    \label{eq:history}
\end{equation} 
The hidden state $h_t$ is leveraged with the action $a_{t-1}$ for both  \textit{posterior inference} and \textit{prior generation}.
The posterior distribution $q_\phi(z_t|o_t, a_{t-1}, h_t)$ projects the high-dimensional observation into the latent space (inference, dashed line in Fig.~\ref{fig:rssm}).
Sampling $q_\phi$ provides the compact semantic of the scene from which the agent can make a decision.
Further, $p_\theta(z_t|a_{t-1}, h_t)$ models the prior knowledge of such semantic given the action \cite{lee2019stochastic}. We can model a function that predicts the future semantics by incorporating $p_\theta$ with proper latent dynamics model, and generate new observations (generation, solid line in Fig.~\ref{fig:rssm}).

GATSBI further encodes the spatio-temporal context by factorizing the latent embedding $z_t$ into the background, agent, and objects.
The history $h_t$ is factorized accordingly to represent the entity-wise states, and we train dedicated LSTMs in Eq.~(\ref{eq:history}) for each posterior-prior sampling of the individual entities.
This way GATSBI maintains the spatio-temporal consistency of different entities.

In addition, we guarantee a comparable contribution of the action to the latent dynamics by enhancing its dimension.
Since the action as a raw vector is relatively low-dimensional compared to the observation, we increase the dimension of the action with a multi-layer perceptron $\hat{a}_t=\mathrm{MLP}(a_t)$.
In contrast to the entity-wise history $h_t$, $\hat{a}_t$ is shared across different modules of GATSBI. 
During the sampling process, the action $a_{t}$ plays a key role in identifying the scene entities and modeling the interaction among them. 

In summary, GATSBI is a recurrent-SSM that samples a disentangled $z_t$ conditioned on $a_{t-1}$ and entity-wise $h_t$. 
In the following, we further explain the action-conditioned entity-wise decomposition (Sec.~\ref{subsec:entity}) and the interaction dynamics between them (Sec.~\ref{subsec:interaction}).

\subsection{Entity-wise Decomposition}
\label{subsec:entity}
\input{fig/schemes/01_overview} 

GATSBI disentangles different entities from the observation sequence and models the interaction between them.
A similar goal has been achieved using attention-based object discovery~\cite{kosiorek2018sequential,stelzner2019faster,crawford2019spatially,crawford2019exploiting,stanic2019r,eslami2016attend}, but they can only represent passive interactions among small objects.
Specifically, they divide the frame into a coarse grid and individual objects are assigned into one of the cells.
However, when the agents are actively interacting with and manipulating objects within the scene, the motions of agents cannot be constrained within the size of a cell.
GATSBI is explicitly designed to locate an active agent in an unsupervised fashion, which appears in diverse motion and shape.

At a high-level, GATSBI decomposes the entities within the observation in three steps as shown in Fig.~\ref{fig:overview}.
First, the \emph{mixture module} acquires latent variables 
that embed the Gaussian mixture model (GMM) of the static background and the active agent.
Next, one of the mixture modes is specified as an agent by the \emph{keypoint module}, whereas the remaining modes are specified as the background.
The keypoint module detects dynamic features in observation, where the movement is highly correlated with the action of the agent.
In the meantime, the \emph{object module} discovers passive scene entities adapting attention-based object discovery \cite{lin2020improving}.
The resulting entities are the active agent, static backgrounds, and the passive objects. 
Finally, the \emph{interaction module} constructs the agent-centric interaction graph with the decomposed entities, and updates the hidden state of the object properties. 
This makes GATSBI accurately reflect the complex interactions caused by the agent.
%
\paragraph{Mixture Module.}
\input{fig/schemes/02_mixture_module} 
The \acrshort{gmm}-based representation learning \cite{burgess2019monet, engelcke2019genesis, greff2019multi} is one way to extract separate entities in the latent representation given the observation $o$.
In contrast to the standard latent representation $z$ of VAE, it assumes that there exist $K$ entities in the scene, and each entity is embedded into separate latent variables $z_k,~k=1,\ldots,K$ that follow a Gaussian distribution.
Therefore the overall distribution is represented as the mixture of $K$ Gaussians.

We handle the structure and the appearance of individual components separately, and this information should be consistently propagated over a time sequence.  
As shown in Fig.~\ref{fig:mix_module}, omitting the time index $t$, GATSBI factorizes the latent variable for each entity $z_k$ into a mask $z^m_k$ and the corresponding component $z^c_k$. 
The observation likelihood $p_\theta(\mu^{\mathrm{mix}}|z^m_{1:K}, z^c_{1:K})$ conditioned on these is formulated as
\begin{equation}
    \label{eq:genesis}
        p_\theta(\mu^{\mathrm{mix}}|z^m_{1:K}, z^c_{1:K}) = \sum\nolimits_{k=1}^K \pi_\theta(z_k^m) p_\theta(o_k|z_k^c).
\end{equation}
For $k$-th entity, the latent variables for mask $z^m_k$ generate the observation mask of $M$ pixels in the image $\pi_\theta(z_k^m) \in \left[0, 1\right]^M$ whereas $z^c_k$ encodes the component appearance and generates the observation $p_\theta(o_k|z_k^c)$.
The mask variable $z^m_{1:K}$ is formulated such that the occupancy of individual scene entities are decided sequentially, i.e., $\pi_\theta(z^m_{1:K})=\prod_{k=1}^K \pi_\theta(z_k^m|z_{1:k-1}^m)$.
Then $z^c_{1:K}$ is conditioned on the mask $z^m_{1:K}$.
This makes $z^m_k$ first determine how much portion each entity $k$ contributes to $o$ then $z^c_k$ determine how each component looks like.

As mentioned, the spatial decomposition is temporally extended where the entity-wise history $h_{t,k}^m$ and $h_{t,k}^c$ follow the update rule defined as Eq.~(\ref{eq:history}).
At each time step $t$, we condition the sampling of latent variables of the first mask on the enhanced action from the previous time step $\hat{a}_{t-1}$ as well as its own history $h^m_{t,k=1}$,
\begin{equation}
    z^m_{t,1} \sim q_{\phi}(z^m_{t,k=1}|o_t, \hat{a}_{t-1}, h^m_{t,k=1}).
    \label{eq:robot}
\end{equation}
We optimize $q_{\phi}$ and $p_{\theta}$ with the ELBO objective in Eq.~(\ref{eq:elbo_1}). In this way, the posterior network $q_{\phi}$ learns the latent transition from $z^m_{t-1,k}$ to $z^m_{t,k}$ that is induced by $a_{t-1}$. 
In addition, with the sequential inference of the mask latent variables, conditioning on $z^m_{t,1}$ transfers the effect of enhanced action for all modes of the mixture model.
Therefore, action-conditioning effectively increases the correlation between the action and the masks, and eventually coordinates the motion of the agent with the temporal change of the masks.
The equations for the full sampling process and the objective of the mixture module are included in Sec.A.1 of the supplementary material.

With the limited number of modes for the Gaussian mixture,
objects in the observation are less prone to be captured by the mixture module.
The weighted sum of components constitutes a reconstruction of the scene where only the agent and background entities exist
$\mu_t^\mathrm{mix} = \sum_{k=1}^{K} \pi_{t, k}o_{t, k}$.
As only the agent and the backgrounds forms $\mu_t^\mathrm{mix}$, we can find the salient feature which solely consists the objects $o_t - \mu_t^\mathrm{mix}$. We use this for better object discovery. 

\paragraph{Keypoint Module.}
\input{fig/schemes/03_keypoint_module}
Even though the mixture module extracts the spatial layout of different entities, it is not trivial to assign a specific index of modes $k$ for the agent under general visual configuration.
In the keypoint module, we utilize a swarm of $N$ object keypoints \cite{minderer2019unsupervised} to describe the morphology of the agent and also represent the implication of their motions.

Fig.~\ref{fig:kypt_module} describes how the keypoint module can extract the agent information from the mixture module.
Given observation, the keypoint module detects salient features that actively move in response to the enhanced action $\hat{a}_t$ as a set of keypoints.
The detected keypoints are aggregated to construct a keypoint map $\gamma_t$, from which we can compare and select the matching index $k$ of the mixture mode. The details for finding the index are described in Sec.~A.2 of the supplementary material.

More importantly, we modify the training objective in \cite{minderer2019unsupervised} as
\begin{equation}
        \label{eq:kl_kypt}
       \kl{q_\phi(z^r_t|o_t, h^r_t, \hat{a}_{t-1})}{p_\theta(z^r_t|h^r_t, \hat{a}_{t-1})} + \|\gamma_t - \pi_{t}^r\|.
\end{equation}
The former term is the KL-divergence from ELBO in Eq.~(\ref{eq:elbo_1}) conditioned on the history of the keypoints $h^{r}_t$ and the enhanced action $\hat{a}_{t-1}$. 
The latter term represents the pixel-wise $l_2$ distance between the aggregated keypoint map $\gamma_t$ and the mask of the robot agent  $\pi^r_{t}=\pi_\theta(z^{m}_{t,k=r})$ with the index $k=r$ specified for the agent.
\paragraph{Object Module.}
The object module adapts the attention-based object discovery by G-SWM \cite{lin2020improving} to find small objects that could not be captured by the mixture module. 
In addition, the object module can discover the rich attributes of individual components as well as their relational context.
For the completeness of the discussion, we briefly introduce the formulation.

The input scene is first divided into coarse grid cells \cite{eslami2016attend}.
For each $(u,v)$-th cell of the 2D grid, a list of latent attributes are specified as $z_{(u,v)}=(z_{(u,v)}^\mathrm{pres}, z_{(u,v)}^\mathrm{where},z_{(u,v)}^\mathrm{what},\ldots)$.
Each of the latent variables represents: the likelihood for its existence; position in the image space; its appearance; and optional other features~\cite{kossen2019structured, crawford2019spatially}.
The dynamic history $h^o_t$ of the latent vectors $z^o_t=\{z_{(u,v)}\}_t$ is condensed with a recurrent-SSM as other modules such that the module maintains the temporal consistency.
The explicit representation of the latent vectors $z^o_t$ enables probabilistic encoding of the various interactions in the state $h^o_t$. 
The information is accumulated in $h^o_{t}$ using a fully-connected graph neural network~\cite{kossen2019structured, lin2020improving}, whose nodes represent the discovered entities, and the edges encode the dynamic interaction between them.
We further extend the approach and posit our agent-centric object interaction.
\subsection{Interaction}
\label{subsec:interaction}

The interaction module models the agent-centric interaction and can generate physically plausible future frames.
After the entity-wise decomposition, GATSBI can extract information of the active agent $z^r_t, h^r_t$ and $I$ passive objects
$z^o_{t,i}, h^o_{t,i}, i \in I$.
The graph-based interaction in  \cite{kossen2019structured, lin2020improving} encodes the interaction dynamics of object $i$ using the object feature $u_{t,i}$,
\begin{equation}
    \label{eq:interaction_GNN}
    \tilde{\mathcal{I}}_{t,i}=\sum\nolimits_{j \neq i}f^o(u_{t,i}, u_{t,j}).
\end{equation}

The interaction module of GATSBI extends the above formulation with two modifications.
First, we confine the physical interaction only among $k$ nearest neighbors, instead of the fully-connected graph in Eq.~(\ref{eq:interaction_GNN}).
By focusing on the entities in close proximity, we greatly reduce the number of edges in the graph.
The reduced formulation not only allows the network to handle a larger number of objects, but also enhances the prediction accuracy as shown in the experimental results.

Second, GATSBI can model the interactions considering the spatio-temporal context, and separately handle the active, passive, and static components of other entities.
This is the immediate benefit from the entity-wise decomposition in Sec.~\ref{subsec:entity} and successfully modeling the acting agent within the scene.
The spatial component uses the latent embedding of the object $z_{t,i}^o$ and the surrounding observation, which is obtained by cropping the non-object observation $\mu^\mathrm{mix}_t$ near the object.
Recall that $\mu^{\mathrm{mix}}_t$ is reconstructed from the mixture module and corresponds to the scene without objects.
The temporal aspect of an interaction is calculated along object feature $u_{t,i}^o$ and agent feature $u^r_t$. 
Similar to the object feature in  \cite{kossen2019structured, lin2020improving}, the agent dynamics $u^r_t$ is modeled from the latent variable of the agent $z^r_t$ and its history $h^r_t$.

The total interaction $\mathcal{I}_{t,i}$ upon the object $i$ is 
\begin{equation}
       \label{eq:interaction_gatsbi}
    \sum\nolimits_{j\in\mathcal{N}(i)} f^o(u_{t,i}^o, u_{t,j}^o)  +f^s(\mu^{\mathrm{mix}}_t,u_{t,i}^o) 
    +f^{t}(u_t^r,u_{t,i}^o).
\end{equation}
Here $f^o, f^s$, and $f^t$ are neural networks that encode different interactions: $f^o$ extracts passive interaction among objects included in $\mathcal{N}(i)$, the $k$-nearest-neighbor objects, while $f^t$ encodes the response to the movement of the agent.
Lastly, $f^s$ takes only positional information into account. 
The state of each object $i$ is updated with the aggregated dynamics as $h^o_{t+1,i} = \mathrm{LSTM}(\mathcal{I}_{t,i},h^o_{t,i})$.
As demonstrated in Sec.~\ref{sec:Experiments}, our unsupervised formulation accurately predicts the physical contact between the agent and multiple objects, and learns reasonable consequences to interactions.

\input{fig/samples/00_spatial_decomp}

%% file: fig/schemes/00_background.tex
\begin{figure}[!th]
    \vspace{-0.24cm}
    \centering
    \includegraphics[scale=0.16]{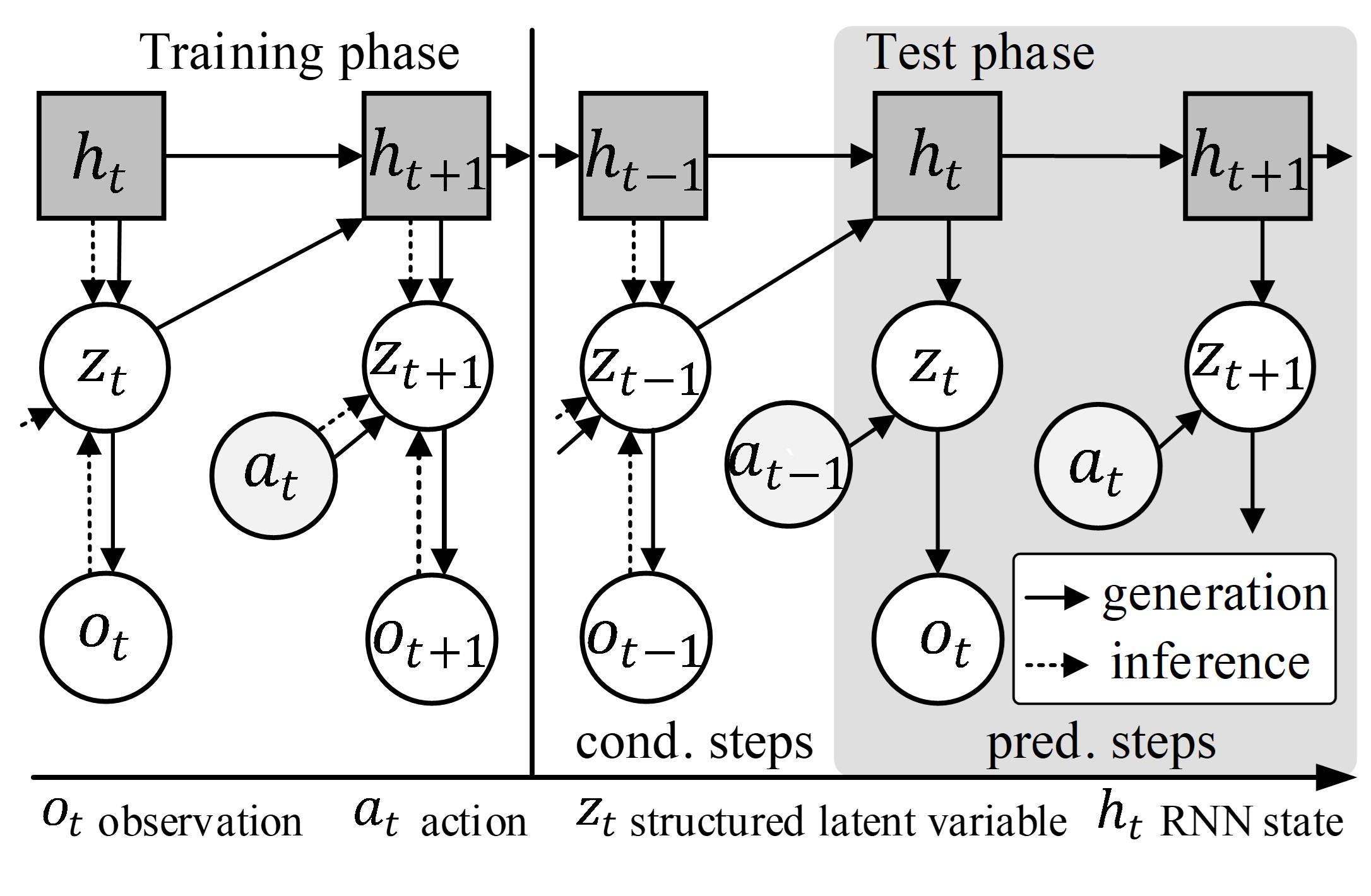}
    \vspace{-0.2cm}
    \caption{A probabilistic graphical model of GATSBI. Left: in the training phase, a set of structured latent variables $z_t$ is inferred (dashed lines) by leveraging recurrent states $h_t$ and observation $o_t$. Right: after updating $h_t$ from $z_t$ conditioned on observations for a few steps, GATSBI consecutively generates (solid lines) the future observations by leveraging recurrent states.}
    \label{fig:rssm}
 \vspace{-0.5cm}
\end{figure}

%% file: fig/schemes/01_overview.tex
\begin{figure}[!th]
    \vspace{-0.37cm}
    \centering
    \includegraphics[scale=0.13]{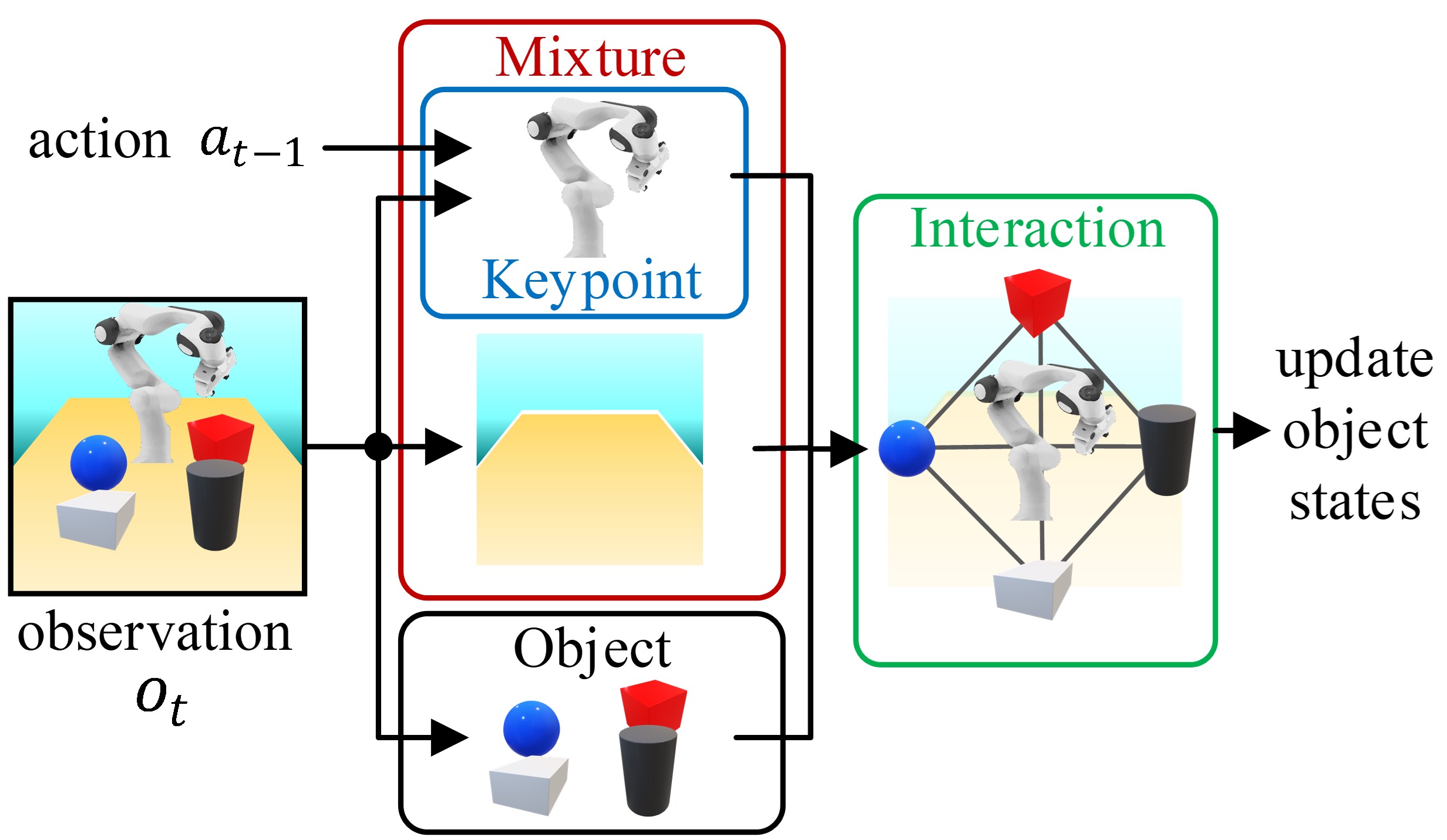}
    \caption{
    An overall scheme of GATSBI. 
    \textit{mixture module} extracts large components leaving the objects. \textit{keypoint module} specifies the agent from the mixture and the remaining entities are assigned as background. 
    \textit{Object module} passes the objects into \textit{interaction module} where a GNN updates the state of objects.
    } 
    \label{fig:overview}
    \vspace{-0.12cm}
\end{figure}

%% file: fig/schemes/02_mixture_module.tex
\begin{figure}[!th]
    \vspace{-0.7cm}
    \centering
    \includegraphics[scale=0.105]{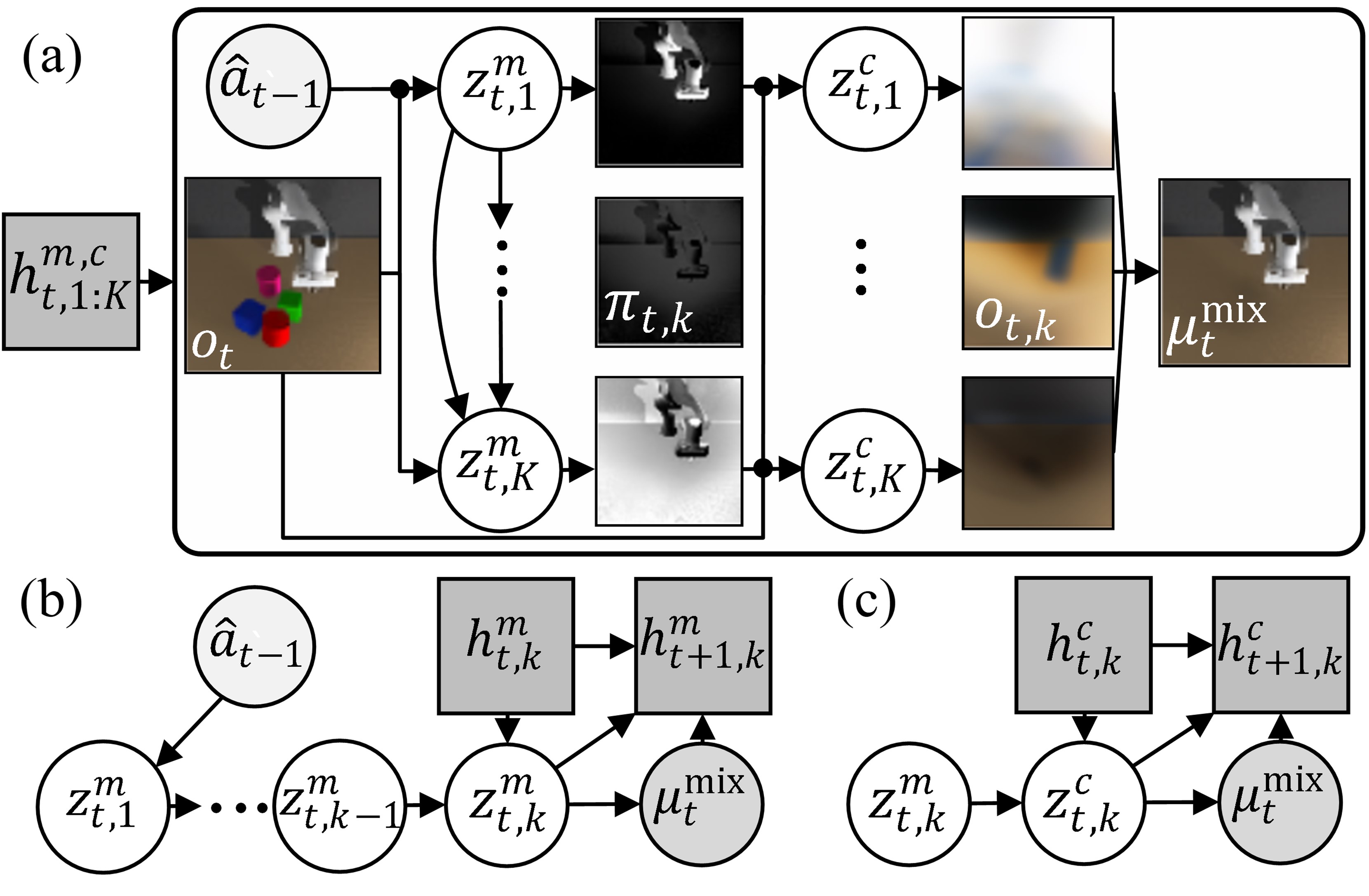}
    \vspace{-0.05cm}
    \caption{
    Spatio-temporal GMM. (a): Conditioning on the recurrent states $h^{m,c}_{t,k}$ and the action of the agent, the mixture module spatially decomposes observation $o_t$ into $K$ individual latent variables $z^{m,c}_{t,k}$ that comprise the mixture $\mu^{\mathrm{mix}}_t$. (b): The recurrent states of each mask variable is temporally updated $h^{m}_{t,k} \rightarrow h^{m}_{t+1,k}$ from autoregressive $z^{m}_{t,k}$ and $\mu^{\mathrm{mix}}_t$, (c): while temporal update of each component variable $h^{c}_{t,k} \rightarrow h^{c}_{t+1,k}$ is done by $z^{c}_{t,k}$ and $\mu^{\mathrm{mix}}_t$.}
    \label{fig:mix_module}
    \vspace{-0.2cm}
\end{figure}

%% file: fig/schemes/03_keypoint_module.tex
\begin{figure}[!th]
    \vspace{-0.75cm}
    \centering
    \includegraphics[scale=0.115]{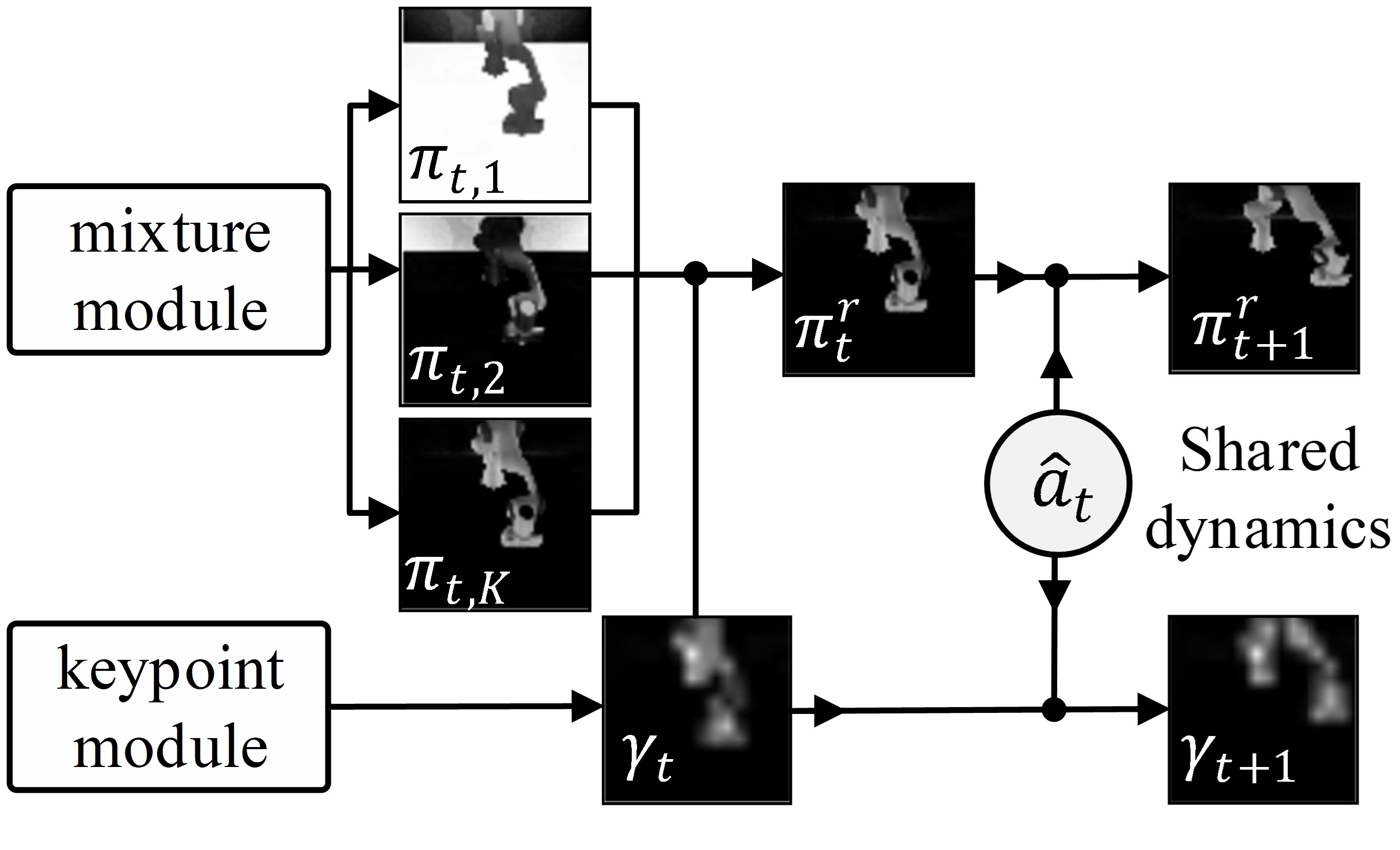}
    \vspace{-0.1cm}
    \caption{Keypoint module. By comparing against the keypoint map $\gamma_t$, we find the index of the agent mask, and fine-tune it to segment out the exact morphology of the agent.
    The dynamics of keypoints and the mask of the agent are shared through enhanced action output.
    }
    \label{fig:kypt_module}
    \vspace{-0.3cm}
\end{figure}

%% file: fig/samples/00_spatial_decomp.tex
\begin{figure*}[hbt!]
    \centering
    \includegraphics[scale=0.295]{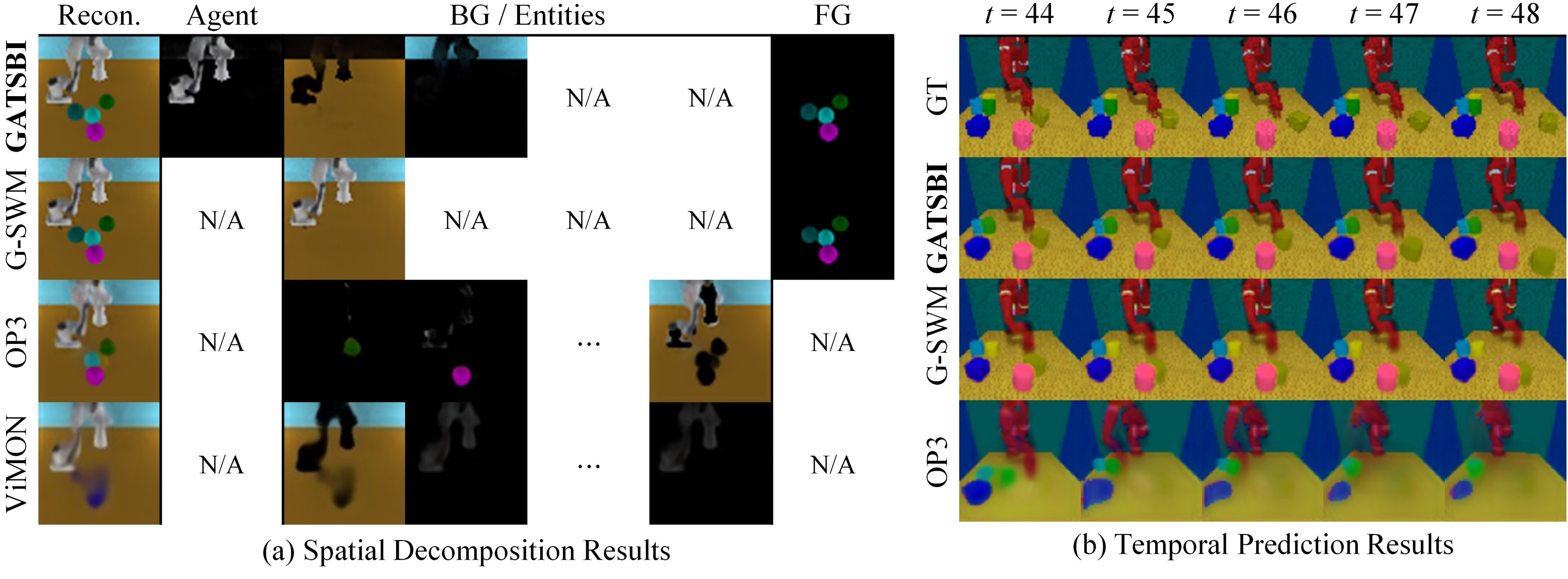}
    \caption{Spatial decomposition and temporal prediction results of GATSBI. (a): For ROLL dataset, GATSBI decomposes a scene into the agent, background, and objects. G-SWM disentangles scene into the background and the objects. Both OP3 and ViMON do not capture explicit scene entities. (b): Long-term prediction results of scenario with a difference of Cartesian pose defined as action. GATSBI predicts the long-term trajectory of agent and its interaction. Prediction of G-SWM is relatively inaccurate, and OP3 loses track of the agent.}
    \label{fig:exp_decomp}
    \vspace{-0.3cm}
\end{figure*}

%% file: 04_experiments.tex
\section{Experiments}
\label{sec:Experiments}
\newacronym{fvd}{FVD}{Fr\'echet video distance}
\newacronym{psnr}{PSNR}{peak signal-to-noise ratios}
\newacronym{ssim}{SSIM}{structured similarity index}
\newacronym{lpips}{LPIPS}{learned perceptual image patch similarity}
\newacronym{mse}{MSE}{mean squared error}
\newacronym{cosim}{COSIM}{cosine similarity}

We evaluate the performance of extracted representation on four synthetic datasets, namely ROLL, PUSH1, PUSH2, and BALLS using physics-based robot simulators~\cite{rohmer2013coppeliasim, james2020rlbench}.
The first three synthetic datasets involve a variety of interactions of agents under different appearances of background, agent, and object, whereas BALLS dataset contains the interaction sequence of multiple balls.
Additionally, we use BAIR~\cite{ebert2017self} to test our algorithm in a real-world dataset.
The input observation is a video sequence that contains a robot agent interacting with objects, and the action space of the agent is defined as the 7 degree-of-freedom (DoF) joint velocities (6 DoF pose of the end-effector + gripper). 
The code and the dataset are available.\footnote{\href{https://github.com/mch5048/gatsbi}{\texttt{\small https://github.com/mch5048/gatsbi}}}
GATSBI is compared against the state-of-the-arts in the structured scene prediction: G-SWM~\cite{lin2020improving}, OP3~\cite{veerapaneni2019entity}, and concurrent work ViMON~\cite{zablotskaia2020unsupervised}.
We first show the results of spatial decomposition in Sec.~\ref{subsec:spatial_decomp} then examine the spatio-temporal prediction in Sec.~\ref{subsec:spatio_temp}.
Finally, the design choices are verified with the ablation study in Sec.~\ref{subsec:ablation_inter}.
Additional experimental results and settings are in the supplementary material.

\subsection{Qualitative Results on Spatial Decomposition} 
\label{subsec:spatial_decomp}
The spatial decomposition is verified by a precise segmentation of the agent, background, and objects. Fig.~\ref{fig:exp_decomp}~(a) shows the spatial decomposition result with the ROLL dataset.
OP3 and ViMON decompose the scene into different mixture modes without knowledge about different entities.
OP3 assigned the robot agent into several slots, and ViMON failed to separate the object entities, which shows an inherent problem of GMM-based approaches \cite{lin2020space}. 
GATSBI overcomes this limitation of the GMM-based approaches by combining with attention-based object discovery, and successfully represents both amorphous shape and small entities. 
G-SWM utilizes attention-based object discovery to detect multiple entities of foreground objects but fails to represent a complex environment with the background because they use a uni-modal Gaussian representation.
While all previous works do not model the agent layer, GATSBI is designed to explicitly decompose the scene into background, agent, and objects.

%
\subsection{Agent-centric Spatio-temporal Interaction}
\label{subsec:spatio_temp}
The precise spatial decomposition of GATSBI plays an essential role to make a physically plausible prediction in response to the agent.
We test the performance of the video-prediction task.
Given the initial five frames of a video sequence, the task is to predict the subsequent frames.
For a fair comparison, previous works are modified to observe action sequence in the latent dynamics model.
For G-SWM, which adopts a recurrent-SSM as GATSBI, we additionally augment its background latent dynamics with the input action, as the background slot is assumed to contain information related to agent movements.
We use the configuration of OP3 that uses the action sequence to train for BAIR cloth manipulation dataset \cite{ebert2018robustness} and ViMON is also modified to adopt the action in the latent dynamics.

Fig.~\ref{fig:exp_spatiotemp_panpu} shows a subset of frames predicted after observing the first five frames of PUSH1 dataset.
As expected, GATSBI generates the agent-object interaction sequence that is nearly identical to the ground truth.
G-SWM predicts a similar configuration of the agent, but the resulting movement of the object is not correctly predicted. 
OP3 generates slightly degraded robot agent configurations. 
\input{fig/samples/01_spatiotemp_panda}
The agent in ViMON is approximately similar, but the shape is blurry and not exact. 
The results imply that the segmentation of the agent and the agent-centric interaction contribute to accurate prediction of both the trajectory of the agent and the consequences of physical interaction. 

The contribution of agent-centric representation of GATSBI is more prominent when tested with the real dataset, Fig.~\ref{fig:exp_spatiotemp_bair1}. Even though the motion of the agent in BAIR dataset is much more stochastic than the synthetic datasets, GATSBI robustly predicts the noisy movement of the agent. 
Since GATSBI adopts the object discovery module from G-SWM, the reconstructions of foreground objects of the two models are nearly identical.
However, G-SWM fails to predict the trajectory of the agent as the agent and action information is mixed in the background slot whereas GATSBI dedicates a separate layer for the agent.  
OP3 makes a relatively accurate prediction on the trajectory of the agent, but fails to capture the meaningful context of the scene, and ViMON totally fails to generate meaningful temporal context.

The graphs on the right side of Fig.~\ref{fig:exp_spatiotemp_panpu} and~\ref{fig:exp_spatiotemp_bair1} present the quantitative evaluation of the video prediction in terms of peak signal-to-noise ratios (PSNR) and learned perceptual image patch similarity (LPIPS) \cite{zhang2018unreasonable}.
PSNR (higher is better) is a widely-used metric for video prediction that aggregates the pixel-wise differences of the predicted frames compared to the ground truth, and LPIPS (lower is better) measures how realistic the predicted frames are.
GATSBI achieves superior performance in terms of both metrics.
We observe that the mixture models of OP3 and ViMON have limited capacity to express detailed visual features and cannot faithfully recover the observation even for the first five frames (shaded in gray) where the ground truth is given.
After the five frames, the system starts to make pure predictions and the performance rapidly deteriorates for all other approaches.
In contrast, the curves for GATSBI are relatively smooth in both PSNR and LPIPS.
It demonstrates that GATSBI leverages the information from observation much more effectively than other methods.
Additional results with all of the datasets are available in the Sec.~E of supplementary material.

\input{tab/01_FVDs}
Table~\ref{tab:exp_fvds} summarizes the performance with all four datasets measured with Fr\'echet video distance (FVD) \cite{unterthiner2018towards}.
FVD (lower is better) measures the distance in the feature space to reflect the similarity of human perception.
GATSBI correctly models the latent dynamics of the agent and objects, and consistently exhibits superior results in all datasets.
G-SWM can make a relatively precise trajectory prediction on synthetic datasets with the entity-wise decomposition and outperforms OP3 and ViMON by a large margin. However, it fails to model the agent-object interaction.

We further evaluate the GATSBI with PUSH2 dataset which we created with a different agent that moves significantly,  interacting with more objects.
In addition, we create a challenging setting by providing the change of translation and rotation of end-effector. 
The correct action configuration needs to be inferred from the relative action information and the history of agent motions.
Fig.~\ref{fig:exp_decomp}~(b) shows the five consecutive predicted frames with the noticeable agent-object interaction.
The robot agent moves its end-effector and hits the yellow cube. 
GATSBI, with the explicit embedding of the agent dynamics incorporated in the interaction model, predicts the passive movement of the yellow cube.  
In contrast, G-SWM only predicts the motion of the agent and fails to capture the interaction.
\input{fig/samples/03_spatiotemp_bair1}
Lastly, OP3 and ViMON show poor prediction of the agent, and could not propagate the objects through time.
\subsection{Ablation Study on Interaction}
This section provides the ablation study on the interaction module. Further studies on the mixture and keypoint module are included in the Sec.~E.6 of supplementary material.
\label{subsec:ablation_inter}
\vspace{-0.4cm}
\paragraph{Comparison of Interaction Modes.}
\label{parag:intermode}
Here we compare different methods of processing interactions, and demonstrate that the latent information of the agent enhances the performance of video prediction.
\input{tab/02_Intermodes}
First mode considers the interaction of individual objects as G-SWM, but the remaining components are regarded as a static background (INTER1).
The other two methods extract the variables of the agent and
The latent dynamics is incorporated into the interaction graph.
INTER2 encodes the variable of the agent as a localized feature for each object, whereas INTER3 (\emph{ours})  uses it as a global feature of the interaction network.
We provide the detailed implementation of each mode in the Sec. A.3 of supplementary material. 
Table~\ref{tab:exp_intermodes} shows the comparison among the three interaction models in terms of PSNR, LPIPS, and FVD in $95\%$-confidence interval. 
INTER3 performs the best, implying that the agent information provides sufficient constraints on all the objects within the scene.
\vspace{-0.4cm}
\paragraph{Agent-free Object Interactions.}
\label{parag:balls_inter}
Finally, we evaluate the $k$ nearest neighbors search method in the object-object interaction model of GATSBI.
We generate synthetic scenes where multiple objects interact with each other, and test the accuracy of video prediction with scenes.
The synthetic scenes contain interactions of different numbers of balls as shown in the inset of Table~\ref{tab:balls_interaction}.
Table~\ref{tab:balls_interaction} presents the numerical pixel error before and after the interaction among objects. 
The result exhibits that the precision of interaction increases as the number of objects increases and outperforms the original fully-connected graphical model. 
With the sparse graph, the network better captures the physical context between multiple objects.
\input{tab/04_balls_inter} 



%% file: fig/samples/01_spatiotemp_panda.tex
\begin{figure*}[ht!]
    \vspace{-0.3cm}
    \centering
    \includegraphics[scale=0.275]{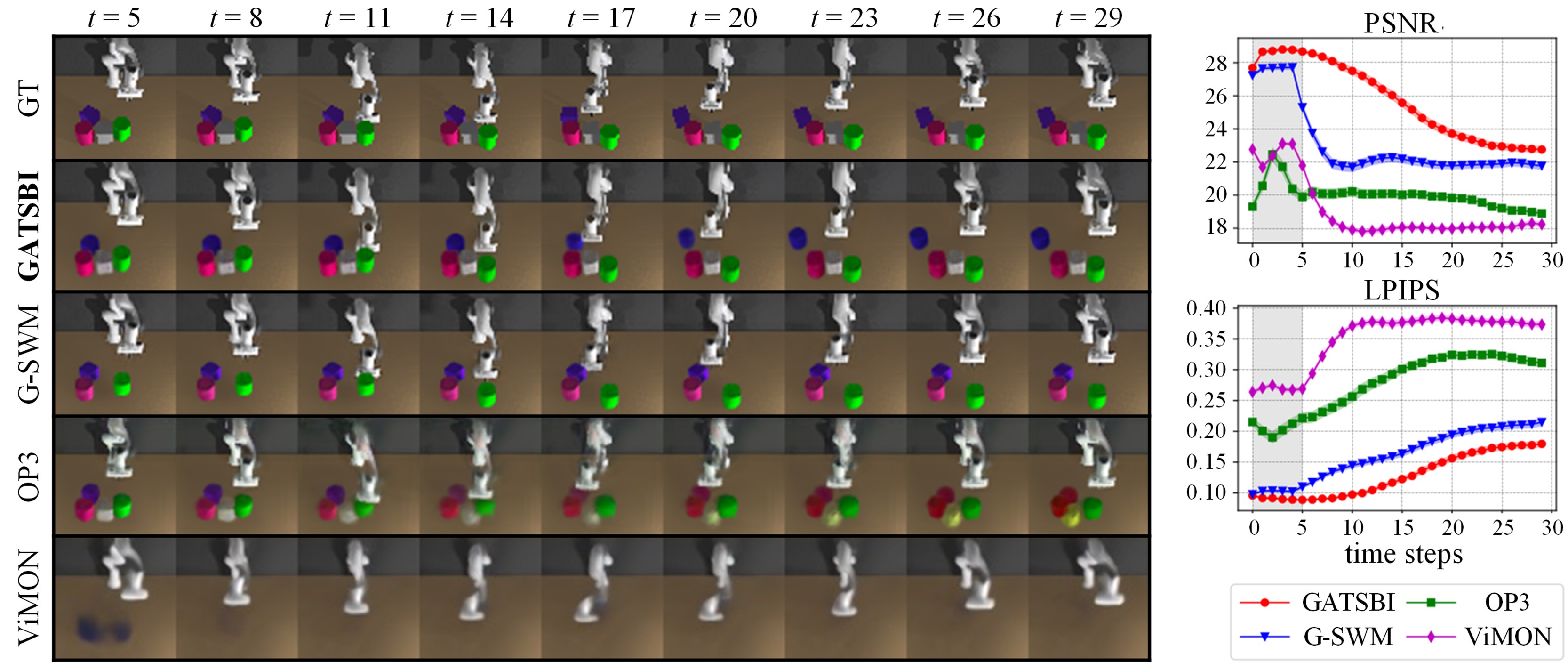}
    \caption{Spatio-temporal prediction results on PUSH1 dataset. Left: the prior generation process results over 25 prediction steps. The figure compares the reconstruction of predicted futures for each method. Right: Quantitative evaluation of predicted video frames. PSNR (\emph{higher is better}) and LPIPS (\emph{lower is better}) are plotted in $95\%$-confidence interval.}
    \vspace{-0.3cm}
    \label{fig:exp_spatiotemp_panpu}
\end{figure*}

%% file: tab/01_FVDs.tex
\begin{table}
    \centering
         \caption{
        \label{tab:exp_fvds}
        FVD (\emph{lower is better}) comparison for all methods on the four robotics dataset. 
        The lower value implies the generated frames are closer to that of ground truth in the feature space. 
        Bold values indicate the best performing method for each dataset. Values inside the parenthesis denote the $95\%$- confidence interval for each setup.
        }
        \vspace{0.2cm}
        \begin{adjustbox}{max width=\columnwidth}
        \begin{tabular}{c|cccc}
         \toprule
        Models          & {ROLL} & {PUSH1}  & {PUSH2} & {BAIR}  \tabularnewline
        \midrule
        \textbf{GATSBI} & \textbf{484.0} (27.57)  &  \textbf{630.4} (37.68)    &  \textbf{859.0} (35.43)  & \textbf{1620} (55.63) \\
        G-SWM           & 627.3 (30.00)  &  910.6  (76.89)    &  1072 (32.92)  & 2603 (121.4) \\
        OP3             & 1025 (39.33)  &  1118 (35.19)    &  2568 (90.01)  & 2904 (128.0) \\
        ViMON           & 1217 (28.98)  &  1620 (58.9)    &  2823 (93.54)  & 3983 (204.9) \\
        \bottomrule
        \end{tabular}
        \end{adjustbox}
        \vspace{-0.6cm}
    \end{table}

%% file: fig/samples/03_spatiotemp_bair1.tex
\begin{figure*}[!t]
    \vspace{-0.3cm}
    \centering
    \includegraphics[scale=0.27]{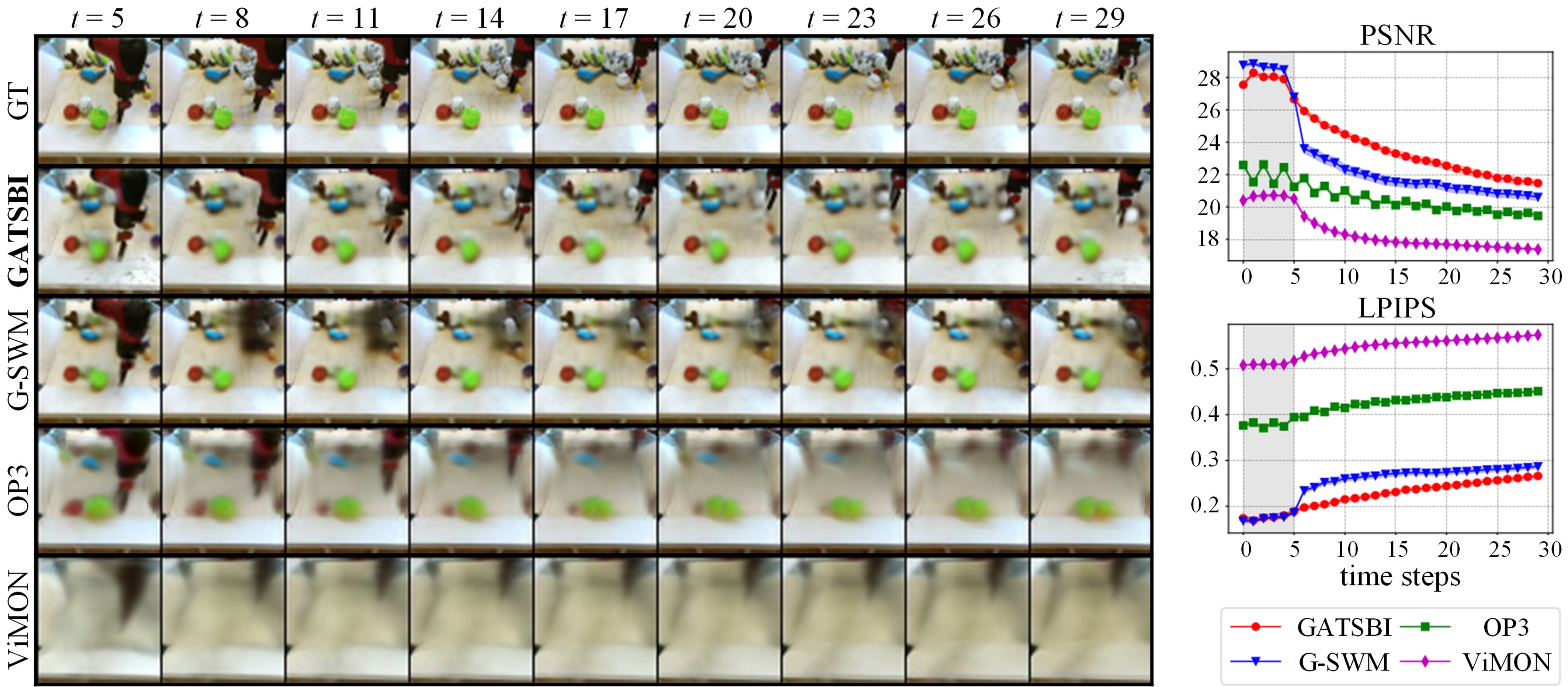}
    \caption{Spatio-temporal prediction results of BAIR dataset. Left: action-conditioned video
    prediction result on real-world robot dataset. Right: PSNR and LPIPS on BAIR dataset. Solid colored mean values are shaded by $95\%$-confidence interval.}
    \label{fig:exp_spatiotemp_bair1}
    \vspace{-0.3cm}
\end{figure*}

%% file: tab/02_Intermodes.tex
\begin{table}
    \centering
         \caption{
        \label{tab:exp_intermodes}
        PSNR (\emph{higher is better}), LPIPS (\emph{lower is better}), FVD comparison among the three interaction modes. 
        }
        \vspace{0.2cm}
        \begin{adjustbox}{max width=\columnwidth}
        \begin{tabular}{l|ccc}
         \toprule
        Mode          & {PSNR} &{LPIPS} & {FVD}  \tabularnewline
        \midrule
        INTER1          &  22.80 (0.2202) & 0.2089 (6.227e-3) & 841.6 (51.17)      \\
        INTER2 &  24.78 (0.1767) & 0.1570 (2.672e-3) & 484.0 (27.575)   \\
        \textbf{INTER3}          &  \textbf{25.44} (0.1765) &\textbf{0.1463} (2.622e-3)     & \textbf{482.5} (22.320)  \\
        \bottomrule
        \end{tabular}
        \end{adjustbox}
        \vspace{-0.6cm}
\end{table}

%% file: tab/04_balls_inter.tex
\vspace{-0.2cm}
\begin{table}[!ht]
    \caption{Average pixel error for different connectivity of interaction graph. FC denotes the fully-connected graph model of G-SWM and KNN ($k$) denotes the $k$ nearest neighbor graph model of GATSBI.}
    \vspace{0.2cm}
    \label{tab:balls_interaction}
    \centering
    \footnotesize
    \begin{tabular}{c|c|ccc}
     \toprule
  \multirow{4}{*}{\includegraphics[scale=0.125]{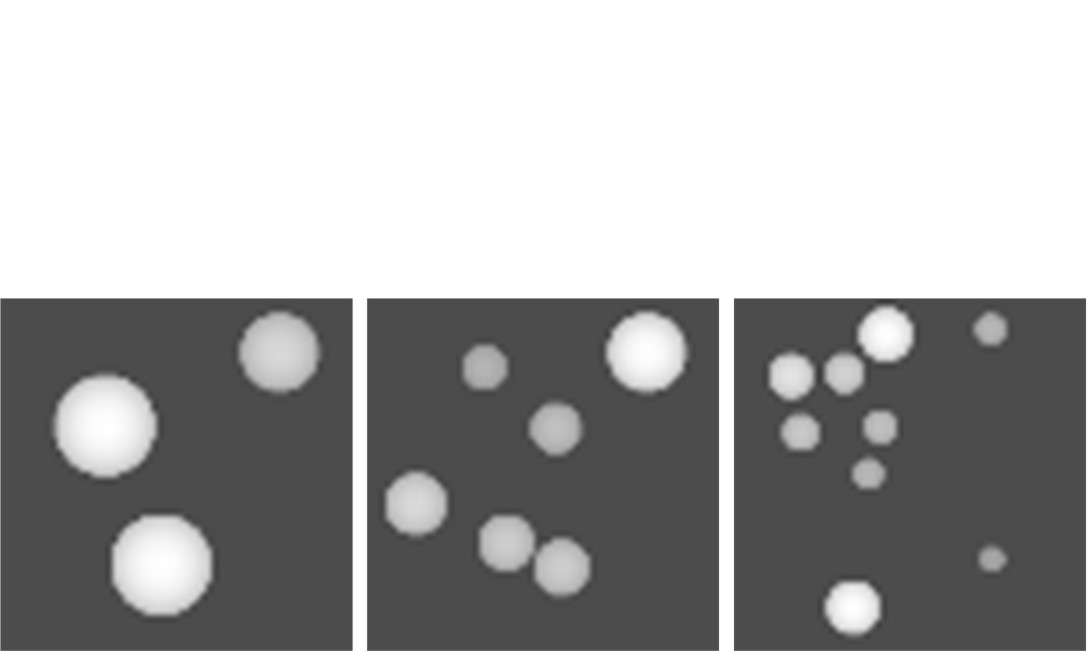}}
       & Method & 3 Balls & 6 Balls & 9 Balls \\
       \midrule
       & FC & \textbf{3.039} & 3.58  & 5.477 \\ 
       & KNN (3) & 3.483 & \textbf{3.374} & 5.975 \\     
       & KNN (5) & - & - & \textbf{3.775} \\ 
     \bottomrule
    \end{tabular}
\vspace{-0.65cm}
\end{table}

%% file: 05_conclusion.tex
\section{Conclusion}
\label{sec:concl}
In this work, we proposed GATSBI, a spatio-temporal scene representation model that decomposes a video observation into an agent, background, and objects. 
With an appropriate representation of the action of the agent, our model reliably predicts the long-term trajectory of the agent as well as the physical interaction between the agent and other objects in the scene. 
The experimental results prove our agent-centric video prediction model can generate physically plausible future frames in various synthetic and real environments. 
Our method excels concurrent state-of-the-art methods both in the qualitative and quantitative results. 
In the future, we will apply GATSBI~to solving vision-based robotics tasks, since our  prediction model can be applied to model-based reinforcement learning.

%% file: 06_acknowledgement.tex
\vspace{-0.3cm}
\section{Acknowledgment}
\label{sec:ackl}
This research was supported by the National Research Foundation of Korea (NRF) grant funded by the Korea government (MSIT) (No. 2020R1C1C1008195) and the National Convergence Research of Scientific Challenges through the National Research Foundation of Korea (NRF) funded by Ministry of Science and ICT (NRF-2020M3F7A1094300).

%% file: supplementary.tex


\section*{\LARGE \bf \centering Supplementary Material for \\GATSBI: Generative Agent-centric Spatio-temporal Object Interaction}

\setcounter{section}{0}
\renewcommand*{\thesection}{\Alph{section}}
\renewcommand*{\thesubsubsection}{\alph{subsubsection}}
\input{01_details_of_modules}
\input{02_objective_functions}
\input{03_detailed_training_test_procedure}

\input{04_details_for_neural_network_models}
\input{05_additional_samples}

%% file: 01_details_of_modules.tex
\vspace{\baselineskip}
\vspace{1cm}
\section{Detailed Explanation of the Modules of GATSBI}
\label{sec:detailed_modules}
In this section, we describe the detailed formulation of each module that constitutes the GATSBI.

\subsection{Mixture Module}
\label{subsec:mixture_details}
As described in Sec.~\ref{parag:mixture} of the paper, we both condition the structured latent variables of mixture model $z^m_{t, 1:K}, z^c_{t, 1:K}$ on the enhanced action of the agent $\hat{a}_{t-1}$ and the history of each latent variable $h^m_{t, 1:K}, h^c_{t, 1:K}$. In detail, $\hat{a}_{t-1}$ is conditioned for all the inference, generation, and reconstruction procedure of $z^m_{t, 1:K}, z^c_{t, 1:K}$. In addition,  
we condition $h^m_{t, 1:K}, h^c_{t, 1:K}$ for both the inference and generation steps as
\begin{equation}
    \label{eq:mixture_condition}
    \begin{aligned}
            &z^m_{t, 1:K} \sim q_\phi(\cdot|o_t, \hat{a}_{t-1}, h^m_{t,1:K}) = q_\phi(z^m_{t, K}|o_t, z^m_{t,K-1}, h^m_{t,k=K})~\cdots~q_\phi(z^m_{t,k=1}|o_t, h^m_{t,k=1}, \hat{a}_{t-1})~~  \cdots \text{inference of}~z^m_{t,1:K} \\
           &z^m_{t, 1:K} \sim p_\theta(\cdot| \hat{a}_{t-1}, h^m_{t,1:K}) = p_\theta(z^m_{t, K}|z^m_{t,K-1}, h^m_{t,k=K})~\cdots~p_\theta(z^m_{t,k=1}|h^m_{t,k=1}, \hat{a}_{t-1})~~ \cdots \text{generation of}~z^m_{t,1:K} \\
            &\pi_{t,1:K} \sim p_\theta(\pi_{t,1:K}| \hat{a}_{t-1}, z^m_{t,1:K})~~ \cdots \text{reconstruction of}~z^m_{t,1:K} \\
            &z^c_{t, 1:K} \sim q_\phi(\cdot|o_t, z^m_{t, 1:K}, \hat{a}_{t-1}, h^c_{t,1:K})~~ \cdots \text{inference of}~z^c_{t,1:K}\\
            &z^c_{t, 1:K} \sim p_\theta(\cdot|z^m_{t, 1:K}, \hat{a}_{t-1}, h^c_{t,1:K})~~ \cdots \text{generation of}~z^c_{t,1:K}\\
            &o_{t,1:K} \sim p_\theta(o_{t,1:K}| \hat{a}_{t-1}, z^c_{t,1:K})~~ \cdots \text{reconstruction of}~z^c_{t,1:K}.
    \end{aligned}
\end{equation}
One thing to note that is $\hat{a}_{t-1}$ is conditioned on the posterior and prior distribution of the first entity $k=1$, in order to share the effect of action between all entities using the autoregressive property. The history $h^m_{t, 1:K}, h^c_{t, 1:K}$ used for each inference and generation step of $z^m_{t, 1:K}, z^c_{t, 1:K}$ are separately updated via Eq.~(\ref{eq:history}) with an individual network for each entity
\begin{equation}
    \label{eq:separate_rnn}
    \begin{aligned}
            &h^{m,c,\mathrm{prior}}_{t,k} = \mathrm{LSTM}^{m,c,\mathrm{prior}}_k(z^{m,c}_{t,k}, \mathrm{CNN}(\mu^{\mathrm{mix}_t}), h^{m,c,\mathrm{prior}}_{t-1,k}) ~~\forall k \\
            &h^{m,c,\mathrm{post}}_{t,k} = \mathrm{LSTM}^{m,c,\mathrm{post}}_k(z^{m,c}_{t,k}, \mathrm{CNN}(\mu^{\mathrm{mix}_t}), h^{m,c,\mathrm{post}}_{t-1,k}) ~~\forall k. \\
    \end{aligned}
\end{equation}
Note that the reconstruction of the mixture of agent and background $\mu^{\mathrm{mix}}_t$ is again to give the LSTM network as an input. In this way, we can compensate the information loss regarding the observation during the two-staged inference of $z^m_{t, 1:K}, z^c_{t, 1:K}$.

Since the mask and the component of the agent and background gradually change (i.e., the agent does not make an abrupt movement.), it is undemanding for the network to learn the temporal difference of the two consecutive observations. Accordingly, we do the residual update of the latent variables \cite{chen2018neural, franceschi2020stochastic}, reshaping the posterior and prior samplings in Eq.~(\ref{eq:mixture_condition}) as
\begin{equation}
    \label{eq:residual}
    \begin{aligned}
            &z^m_{t+1, 1:K} = z^m_{t, 1:K} + \Delta \cdot \MLP^{\mathrm{mask}}_{\mathrm{res}}(\bar{z}^m_{t+1, 1:K}, z^m_{t, 1:K}) \\
            &z^c_{t+1, 1:K} = z^c_{t, 1:K} + \Delta \cdot \MLP^{\mathrm{comp}}_{\mathrm{res}}(\bar{z}^c_{t+1, 1:K}, z^c_{t, 1:K}). \\
    \end{aligned}
\end{equation}
$\MLP^{\mathrm{mask}}_{\mathrm{res}}, \MLP^{\mathrm{comp}}_{\mathrm{res}}$ are MLPs that approximate the quantity of the residual update and $\Delta$ is the scaling factor that corresponds to tha sampling period of the episodes. Since $\bar{z}^m_{t+1, 1:K}, \bar{z}^c_{t+1, 1:K}$ are the original outputs from $q_\phi, p_\theta$. Detailed values of $\Delta$ are reported in the Sec.~\ref{subsec:hyper_nn}. 

\subsection{Keypoint Module}
As described in Sec.~\ref{parag:kypt} of the paper, the keypoint module of GATSBI further extends \cite{minderer2019unsupervised}. From the $N$ $G \times G$-sized feature maps $\bar{g}_{t,n}~n=1,\cdots,N$ that embed the difference of current observation to the first frame $o_t - o_0$, we sum up all $N$ feature maps element-wise and increase the dimension via cubic interpolation
\begin{equation}
    \label{eq:kypt_map}
    \begin{aligned}
            \gamma_t = f^{\mathrm{kypt}}_{\mathrm{map}}(g_t) \in \left[0, 1 \right]^{H \times W},~\text{where}~g_t=    \sigma \left( \sum_{n=1}^N \bar{g}_{t,n} \right) .
    \end{aligned}
\end{equation}
$g_t$ is a sigmoid-activated sum of $N$ keypoint features maps and  $f^{\mathrm{kypt}}_{\mathrm{map}}(g_t)$ is a cubic interpolation layer that outputs the mask $\pi_{t,k}$ sized keypoint map. As the nature of keypoints drives $\gamma_{t}$ to capture the morphology of the agent, we can find the agent mask from the $K$ candidates $\pi_{t,1:K}$ that has the most similar shape to $\gamma_{t}$ as
\begin{equation}
    \label{eq:find_match}
    \begin{aligned}
            k_{\mathrm{agent}} = \underset{k}{\arg\max} \left\| \left(\mathbbm{1}_{\pi_{t,k}^{h,w}>0.5} \right) \cdot \left(\mathbbm{1}_{\gamma_{t}^{h,w}>0.5} \right) \right\|.
    \end{aligned}
\end{equation}
$\mathbbm{1}_{\cdot>0.5}$ is an indicator function that indicates whether each pixel value of $\pi_{t,k}$ and $\gamma_{t}$ is larger than $0.5$. Thus, we choose the index of the agent entity $k_{\mathrm{agent}}$ as whose mask $\pi_{t,k}$ has the most overlapping pixels (of values larger than $0.5$) with $\gamma_{t}$.
Once $k_{\mathrm{agent}}$ is specified, the agent mask $\pi_t^r = \pi_{t,k_{\mathrm{agent}}}$, agent latent variable $z_t^r = z_{t,k_{\mathrm{agent}}}$, and history $h_t^r = h_{t,k_{\mathrm{agent}}}$ are subsequently defined.
\subsection{Interaction Module}
\label{subsec:interaction_details}
The agent that gives causality to the movement of the object entities is conditioned on its action, and the background acts as a global constraint for the movement of object entities. Therefore, these causal relationships should be properly defined so as to the dynamics of passive objects to be affected by their surroundings.

As clarified in Sec.~\ref{subsec:interaction}, we model the interactions among scene entities as the total interaction for each object $i$ in Eq.~(\ref{eq:interaction_gatsbi}). 
First, the k-nearest-neighbor objects $\mathcal{N}(i)$ of object $i$ are found as
\begin{equation}
    \label{eq:knn_search}
    \begin{aligned}
            &\mathcal{N}(i) = \mathrm{KNN}_j(W_{ij}),~\text{where}~W_{ij} = f^{\mathrm{rel}}_{\mathrm{dist}}(u_{t,i}, u_{t,j}).
    \end{aligned}
\end{equation}
$u_{t,i}$ and $u_{t,j}$ are feature of discovered objects and $f^{\mathrm{rel}}_{\mathrm{dist}}$ is a function that computes a weight matrix $W_{ij}$ that embeds pairwise relative distance for each object pair $ij$ as proposed in \cite{lin2020improving}. Subsequently we do the $k$-nearest neighbor search $\mathrm{KNN}_j$ to find a set of $k$ nearest entities $\mathcal{N}(i)$ for each object $i$. Finally, summing up the pairwise interaction features between the object $i$ and $j$ yields the object-object interaction portion of the total interaction in Eq.~(\ref{eq:interaction_gatsbi})
\begin{equation}
    \label{eq:oo_interaction}
    \begin{aligned}
            &e^o_{t,i} = \sum_{j\in \mathcal{N}(i)} f^o(u_{t,i}, u_{t,j}),
    \end{aligned}
\end{equation}
where $f^o$ is a function that extracts the pair-wise feature of interaction between object $i$ and $j$ into the feature vector $e^o_{t,i}$.

Next, $f^s$ in Eq.~(\ref{eq:interaction_gatsbi}) extracts the spatial constraints for each object made by the entities of the mixture model: objects should lie on top of the ground, they should not pass through the walls, and they cannot penetrate the agent. In that regard, the agent for each time step $t$ can be deemed as a static scene entity like other background entities.
Therefore, we crop the neighboring part of each object $i$ in non-object observation $\mu^\mathrm{mix}_t$ using each object latent variable $z^o_{t,i}$ and encode the cropped image as
\begin{equation}
    \label{eq:static_interaction}
    \begin{aligned}
            e^s_{t,i} = f^s(\mu^\mathrm{mix}_t, z_{t,i}).
    \end{aligned}
\end{equation}
$f^s$ is a function that encodes the positional information of each object $i$ and its surroundings into the feature vector $e^s_{t,i}$.

Finally, we define a function that embeds the dynamics of the agent to model the agent-object interaction. As the history of the state-action underlies the dynamics of the agent, we consider both the current state of the mask $z^r_t=z^m_{t,k_{\mathrm{agent}}}$ and the history $h^r_t=h^m_{t,k_{\mathrm{agent}}}$ of the agent mask in modelling the agent dynamics. With the concatenation of the two information $u^r_t = (z^r_t, h^r_t)$, we consider two types of interaction.
First, the INTER2 in Sec.~\ref{parag:intermode} of the paper explicitly pairs the local interaction between the agent and each object $i$ as 
\begin{equation}
    \label{eq:temporal_interaction_local}
    \begin{aligned}
            &u^\mathrm{loc}_{t,i} = f^{\mathrm{loc}}(u^r_t, u_{t,i}) \\
            &W^{\mathrm{loc}}_{t,i} = f^{\mathrm{w}}(u^{\mathrm{pos},r}_t, u^\mathrm{pos}_{t,i})\\
            &e^t_{t,i} = f^{t,\mathrm{loc}}(W^{\mathrm{loc}}_{t,i} \cdot u^\mathrm{loc}_{t,i}). 
    \end{aligned}
\end{equation}
$f^{\mathrm{loc}}$ first encodes the pairwise local feature $u^\mathrm{loc}_{t,i}$ between the agent and object $i$, and $f^w$ formulates the attention weight $W^{\mathrm{loc}}_{t,i}$ for each local interaction. We give $f^w$ the positional information of the agent $u^{\mathrm{pos},r}_t = (\hat{a}_t, h^r_t)$ and each object $u^\mathrm{pos}_{t,i} = (z^\mathrm{where}_{t,i}, h_{t,i})$ as inputs. Consequently, the feature vector $e^t_{t,i}$ that embeds the dynamical interaction between the agent and object $i$ is extracted from the weighted local feature.

The other much simple and experimentally better performing approach denoted as INTER3 in Sec.~\ref{parag:intermode} of the paper considers the information of the agent as a globally shared feature by  
\begin{equation}
    \label{eq:temporal_interaction_global}
    \begin{aligned}
            &e^t_{t,i} = f^{t,\mathrm{glob}}(u^r_t).
    \end{aligned}
\end{equation}
For all experiments of this paper, we use INTER3 model to formulate $e^t_{t,i}$ that contributes to the total interaction in Eq.~(\ref{eq:interaction_gatsbi}) of the main paper.

%% file: 02_objective_functions.tex
\newpage
\section{Objective Function}
\label{sec:objective_function}

In this section, we introduce the full training objective of GATSBI.
\label{subsec:mixture_objective}
As noted in Sec.~\ref{sec:gatsbi}, our model follows the formulation that expands the ELBO in Eq.(\ref{eq:elbo_1}) temporally as 
\begin{equation}
    \label{eq:elbo_ssm}
    \begin{aligned}
    &\E \left[ \log p_\theta(o_{0:T}|a_{0:T}) \right] \geq \E_{z_t \sim q} \biggl[ \sum_{t=0}^{T} \log p_\theta(o_t|z_t) - \kl{q_\phi(z_t|o_{\leq t}, z_{<t}, a_{t-1})}{p_\theta(z_t|z_{<t}, a_{t-1})} \biggr].
    \end{aligned}
\end{equation}
The optimization rule of Eq.~(\ref{eq:elbo_ssm}) is commonly applied to the three modules of GATSBI: mixture module, keypoint module, and object module. We elaborate on the training objectives of the three modules here.

\subsection{Mixture Module}
\label{subsec:elbo_mix}
As the mixture module infers structured latent variable $z^m_{t,1:K}, z^c_{t,1:K}$ as in Eq.~(\ref{eq:genesis}), and the temporal update via RNN requires the encoding of a current observation, we can further develop Eq.~(\ref{eq:elbo_ssm}) as 

\begin{equation}
    \begin{aligned}
        \label{eq:elbo_mix}
        &\E \left[ \log p_\theta(o_{0:T}|a_{0:T}) \right] \geq \E_{z^m_{t, 1:K}, z^c_{t, 1:K} \sim q} \biggl[ \sum_{t=0}^{T} \log p_\theta(o_t|z^m_{t, 1:K}, z^c_{t, 1:K}, \bar{a}_{t-1})\\
        &- \kl{q_\phi(z^m_{t, 1:K}, z^c_{t, 1:K}|\mu^{\mathrm{mix}}_{\leq t}, z^m_{<t, 1:K}, z^c_{<t, 1:K}, \bar{a}_{< t})}{p_\theta(z^m_{t, 1:K}, z^c_{t, 1:K}|\mu^{\mathrm{mix}}_{\leq t}, z^m_{<t, 1:K}, \bar{a}_{< t})} \biggr].
    \end{aligned}
\end{equation}
In the right-hand side of Eq.~(\ref{eq:elbo_mix}), the observation likelihood term is conditioned on the enhanced action $\bar{a}_{t-1}$ of Eq.~(\ref{eq:action}) as well as on both the latent variables of \acrshort{gmm}, since we condition the action of the agent as a conditional-VAE (CVAE) \cite{sohn2015learning} scheme.   
In addition, note that both the posterior and prior are conditioned on the history of the reconstructed Gaussian mixture $\mu^{\mathrm{mix}}_{\leq t}$, which is implemented via the LSTM in Eq.~(\ref{eq:separate_rnn}).

As the latent variables of mixture module are updated by inferring their residuals (Eq.~(\ref{eq:residual})), we add regularization terms $\lambda \left\| \Delta \cdot f^{\mathrm{mask}}_{\mathrm{res}}(\bar{z}^m_{t+1, 1:K}, z^m_{t, 1:K}) \right\| + \lambda \left\| \Delta \cdot f^{\mathrm{comp}}_{\mathrm{res}}(\bar{z}^c_{t+1, 1:K}, z^c_{t, 1:K}) \right\|$
to the objective function so that no drastic change occurs, where $\lambda$ is a hyperparameter that determines the degree of the residual regularization in training phase.
\subsection{Keypoint Module}
\label{subsec:elbo_kypt}
The original keypoint learning method that GATSBI bases on \cite{minderer2019unsupervised} has the objective function of
\begin{equation}
    \label{eq:kypt_orig_obj}
    \begin{aligned}
    \mathcal{L}_{\mathrm{kypt}} = \mathcal{L}_{\mathrm{image}} + \lambda_{\mathrm{sep}}\mathcal{L}_{\mathrm{sep}} + \lambda_{\mathrm{sparse}}\mathcal{L}_{\mathrm{sparse}} + \mathcal{L}_{\mathrm{VRNN}} + \mathcal{L}_{\mathrm{future}}.
    \end{aligned}
\end{equation}
$\mathcal{L}_{\mathrm{image}}$ is a reconstruction loss for the keypoint feature maps,  $\mathcal{L}_{\mathrm{sep}}$ and $\mathcal{L}_{\mathrm{sparse}}$ are terms that penalizes the keypoints being overlapped. In addition, $\mathcal{L}_{\mathrm{VRNN}}$ and $\mathcal{L}_{\mathrm{future}}$ are terms that embeds the latent dynamics of keypoint latent variables $z^k_t$ following the traning scheme of \cite{chung2015recurrent}. 

The original paper \cite{minderer2019unsupervised} emphasized the long-term generation of keypoints rather than the one-step latent dynamics of $z^k_t$ as it generates only the prior samples after conditioning steps. In contrast, GATSBI only requires the keypoint map inference during the training phase as we use keypoint module to achieve three goals:
\textbf{1)} the selection of agent index as Eq.~(\ref{eq:find_match}), \textbf{2)} regression between the keypoint map in Eq.~(\ref{eq:kypt_map}) and the agent mask, \textbf{3)} and the coordination of latent dynamics between the keypoints and the agent. Therefore, we modify the KL-divergence term of $\mathcal{L}_{\mathrm{VRNN}}$ as $\kl{q_\phi(z^k_t|o_t, h^k_t, \hat{a}_{t-1})}{p_\theta(z^k_t|h^k_t, \hat{a}_{t-1})}$ to encourage \textbf{3)} and add an additional loss term $ \|\gamma_t - \pi_{t}^r\|$ to $\mathcal{L}_{\mathrm{kypt}}$ for \textbf{2)}. In this way, the latent dynamics between the keypoints and the agent entity are coordinated through the differentiable enhanced action $\hat{a}_{t-1}$, and the regression term jointly tunes the keypoint map and agent mask to better capture the exact morphology of the agent.

%% file: 03_detailed_training_test_procedure.tex
\newpage
\section{Training and Test Procedure}
\label{sec:train_test_detail}
In this section, we elaborate on the training scheme of GATSBI that we use to acquire the results presented in the main paper and this supplementary material. 

\subsection{Training Scheme}
\label{subsec:train_scheme}
At a high-level, GATSBI needs to train three modules: \emph{mixture module}, \emph{keypoint module}, and \emph{object module}. As the scale of the cost of each module differs, jointly training the three modules requires intricate hyperparameter tuning that regulates the scale of each loss term. Thus, we sequentially train the three modules, considering the characteristics of the three modules.

We first train the keypoint module only for certain iterations, as we need to embed the latent dynamics to the parameters of the action enhance network and make the keypoint map capture a rough morphology of the agent. After the keypoint-only training steps, we jointly train the keypoint module and mixture module except for the object module. The main reason for this is that the scale of loss term of the object module is relatively larger than that of the mixture module as a lot of terms are involved in formulating the object latent variables. Thus, the optimizer prioritizes minimizing the loss term of the object module, which may yield the undesirable outcome that the agent and background entities being broken down into several pieces and captured as an object. Subsequently, we train all three modules jointly for certain iterations and stop the keypoint learning to reduce the computational cost. Note that we adopt the curriculum learning scheme that gradually increases the data sequence for training, which is widely used in this field \cite{kosiorek2018sequential, crawford2019exploiting, veerapaneni2019entity, lin2020improving}.

On top of the training scheduling of the three modules, we fix the $\alpha$ value that weights the pixel-wise mixture between the mixture module and the object module. The total reconstruction of observation of GATSBI  $\mu_t = \mu^{o}_t + (1 - \alpha) \mu^{\mathrm{mix}}_t$ follows the formulation of \cite{lin2020improving, lin2020space}, where the mixing weight $\alpha$ is a trainable value computed by the attributes of object latent variables. We observe that object module in the early stage of training yields near-zero value of $\alpha$ for all robotics datasets we test on, failing to capture the object entities properly since near-zero $\alpha$ makes $\mu_t$ only be described by $\mu^{\mathrm{mix}}_t$. Thus, we fix $\alpha$ value for specified optimization steps at the start of training of the object module to incentivize the object discovery. 

The same training scheme delineated here is also applied to our experiments on G-SWM for a fair comparison, since the official implementation \footnote{\href{https://github.com/zhixuan-lin/G-SWM}{https://github.com/zhixuan-lin/G-SWM}} of G-SWM also suffers from the same issue. For OP3, we train the models following the scheme used to train on BAIR cloth dataset implemented in the official code repository \footnote{\href{https://github.com/jcoreyes/OP3}{https://github.com/jcoreyes/OP3}}. For ViMON, we also used the official training setup \footnote{\href{https://github.com/ecker-lab/object-centric-representation-benchmark}{https://github.com/ecker-lab/object-centric-representation-benchmark}}. The detailed values for the training scheme are introduced in Sec.~\ref{sec:model_parameters}.

For the history of the agent $h_t^r$ which is used for modeling the agent-object interaction as in Eq.~(\ref{eq:temporal_interaction_global}), we choose the recurrent state of the LSTM for the prior distribution $\mathrm{LSTM}^{\mathrm{prior}}_k$ from the separated LSTM in Eq.~(\ref{eq:separate_rnn}). The main reason for this choice is to guarantee the consistency of the history since only the hidden state of the prior distribution is updated during the test phase.

Finally, for the optimization of the objective function of the keypoint module Eq.~(\ref{eq:kl_kypt}) and the mixture module Eq.~(\ref{eq:elbo_mix}), we assume that $\bar{a}_{t=-1}$ that is required for sampling the latent variables at $t=0$ is the same as $\bar{a}_{t=0}$. This assumption is valid as we collected the data offline. For the future work for applying GASTBI to online reinforcement learning, we may train additional networks for $t=0$ that do not require $\bar{a}_{t=-1}$.  

\subsection{Test Scheme}
\label{subsec:test_scheme}
As described in Sec.~\ref{sec:Experiments} of the main paper, all of our experiments on future prediction are conditioned on the first five conditioning frames. In detail, $h^m_{t,1:K}, h^c_{t,1:K}$ of the posterior distribution are only updated for the conditioning time steps to properly update and do the posterior inference of $z^m_{t,1:K}, z^c_{t,1:K}$. These variables then update both the hidden state of the posterior and prior distributions until the conditioning steps end.
During the prediction steps, only the hidden states for the prior distribution are updated with the generated samples $z^m_{t,1:K}, z^c_{t,1:K}$ from $p_{\theta}$. In addition, as noted before, we only use $h^m_{t,1:K}, h^c_{t,1:K}$ for the whole sequence in extracting the feature of the agent that involves in the interaction module.

%% file: 04_details_for_neural_network_models.tex
\newpage
\section{Parameters for Training and Test}
In this section, the detailed values on designing the neural networks, training scheme of GATSBI as well as dataset details are reported.
The format of tables and the notations are borrowed from the supplementary material \cite{lin2020improving} and modified.
\label{sec:model_parameters}
\subsection{Network Details}
For all hidden layer activation, unless otherwise specified, we use CELU \cite{barron2017continuously} activation. For Table.~\ref{tab:arch}, LSTM(\emph{in}, \emph{hid}) denotes the dimension of input feature \emph{in} and the dimension of the history \emph{hid}. We only show the dimension of hidden layers for MLPs.
Note that we differentiate $h^m_{t,k^*}$ which denote the spatial autoregressive state of the mask latent variable from $h^m_{t^*,k}$ which indicate the temporal recurrent state of the mask latent variable.

\label{subsec:hyper_nn}
\begin{table*}[htbp]
    \caption{Network details}
    \centering
    \input{tab/arch/arch}
    \label{tab:arch}
\end{table*}
\twocolumn
\begin{table}[htbp]
    \caption{$\CNN^{\mathrm{post}}_{\mathrm{enc}}$. GN(\emph{n}) denotes group normalization.}
    \centering
    \input{tab/networks/generic/01_obs_encoder}
    \label{tab:post_encoder}
\end{table}
\begin{table}[htbp]
    \centering
    \caption{$\CNN^\mathrm{mask}_{\mathrm{dec}}$. SubConv(\emph{n}) denotes sub-pixel convolution adopted from \cite{lin2020space}.}
    \input{tab/networks/mixture/mask/01_mask_decoder}
    \label{tab:mask_decoder}
\end{table}
\begin{table}[htbp]
    \centering
    \caption{$\CNN^{\mathrm{comp}}_{\mathrm{enc}}$. BN(\emph{n}) denotes batch normalization.}
    \input{tab/networks/mixture/comp/01_comp_encoder}
    \label{tab:comp_encoder}
\end{table}
\begin{table}[htbp]
    \centering
    \caption{$\CNN^{\mathrm{comp}}_{\mathrm{dec}}$}
    \input{tab/networks/mixture/comp/02_comp_decoder}
    \label{tab:comp_decoder}
\end{table}

\paragraph{Keypoint Module and Object Module.}
Unless otherwise, we make an additional description, we use the same neural network architecture as proposed in the original papers for each keypoint module and object module \cite{minderer2019unsupervised, lin2020improving}. 

\newpage

\subsection{Hyperparameter Settings}
\label{subsec:hyper_training}
\begin{table}[htbp]
\caption{General hyperparameters}
    \centering
    \input{tab/hyperparam/hyper}
    \label{tab:hyper}
\end{table}
\subsection{Dataset Explanation}
\label{subsec:dataset_detail}
In this section, we introduce the characteristic of each dataset and the dataset-specific hyperparameters.
\paragraph{ROLL dataset.} It is a dataset of an environment where a $7$ degree of freedom robot agent randomly hits low-frictional balls on a low-frictional surface. Objects collide under nearly an ideal elastic collision condition, a billiard-like environment. It has $5,000$ episodes for training and $1,000$ episodes for the test, each episode having a length of $40$ frames of sampling period about $150\mathrm{ms}$. It has a relatively deterministic trajectory of the agent over the whole episode, while the movement of the object is not. Table~\ref{tab:hyper_rob_obj} reports the detailed hyperparameter setting for ROLL dataset.
\begin{table}[htbp]
\caption{Hyperparameters for ROLL dataset.}
    \centering
    \input{tab/hyperparam/dataset/01_rob_obj_hyper}
    \label{tab:hyper_rob_obj}
\end{table}
\paragraph{PUSH1 dataset.} PUSH1 dataset has a richer interaction between the agent and the objects, where the objects are pushed by the agent on a low-frictional surface. Different from ROLL dataset, the objects tend to stand still, thus the collisions among the scene entities occur frequently. It has $1,700$ episodes for training and $300$ episodes for the test, each episode having a length of $40$ frames of sampling period about $200\mathrm{ms}$. The agent in PUSH1 dataset moves along various trajectories, thus making diverse interactions. Table~\ref{tab:hyper_panda} reports the detailed hyperparameter setting for PUSH1 dataset.
\begin{table}[htbp]
\caption{Hyperparameters for PUSH1 dataset.}
    \centering
    \input{tab/hyperparam/dataset/02_panda_hyper}
    \label{tab:hyper_panda}
\end{table}
\paragraph{PUSH2 dataset.} In the PUSH2 dataset, a robot agent that is different from the previous two datasets interacts with objects. Further, the number of objects is increased and the randomness of the motion of the agent has also increased. What is more challenging is that we change the action space of the agent to the Cartesian translation with a minute change in rotation. With this change, our model should leverage the history of the agent trajectory better since translation and change in rotation do not include the information of the current state of the agent. It has $5,500$ episodes for training and $2,000$ episodes for the test, each episode having a length of $84$ frames of sampling period about $100\mathrm{ms}$. Table~\ref{tab:hyper_sawyer} reports the detailed hyperparameter setting for PUSH2 dataset.
\begin{table}[htbp]
\caption{Hyperparameters for PUSH2 dataset.}
    \centering
    \input{tab/hyperparam/dataset/03_sawyer_hyper}
    \label{tab:hyper_sawyer}
\end{table}
\paragraph{BAIR dataset.} The BAIR push dataset is proposed in \cite{ebert2017self}, where a 7 degree of freedom robot arm hovers over a tray with many objects of various shapes are put. The action space for the trajectory is $3$ dimensional Cartesian space translation. This dataset has the most stochastic motion of the agent among all datasets we test on, and the shape of each object is relatively complex to our synthetic datasets. We split the original data into about $43,300$ episodes for the training and $256$ episodes for the test. Table~\ref{tab:hyper_bair} reports the detailed hyperparameter setting for BAIR dataset.
\begin{table}[htbp]
\caption{Hyperparameters for BAIR dataset.}
    \centering
    \input{tab/hyperparam/dataset/04_bair_hyper}
    \label{tab:hyper_bair}
\end{table}
\paragraph{BALLS dataset} The BALLS dataset consists of video clips of multiple balls of different sizes and speeds moving dynamically. Balls not only hit the rim but also collided with each other, and the conservation of momentum is assumed. Each frame is rendered in the same way as the depth image, and the dataset differs from the existing bouncing ball datasets in that occlusions are completely dependent on the $3$D geometries of balls. The dataset contains $5,000$ episodes for training and $200$ episodes are assigned to the test set.

%% file: tab/arch/arch.tex
\renewcommand{\arraystretch}{1.2}{
\begin{center}
\begin{tabular}{lll}
\toprule
 Functionality & Notation & Architecture \\
\midrule
\textbf{Mixture Module} \\
Encode $o_t$ into $e_{t}$ & $\CNN^\mathrm{post}_{\mathrm{enc}}$ &  Refer to Table~\ref{tab:post_encoder}.\\
Enhance raw action $a_t$ to $\bar{a}_t$ & $\MLP^\mathrm{act}_{\mathrm{enh}}$ & [64, 64]\\
Encode $[e_{t}, h^m_{t^*,k}, (\bar{a}_{t-1})]$ into $e^{m,\mathrm{post}}_{t,k}$ & $\MLP^{\mathrm{post}}_{\mathrm{mask}}$ &  [128, 128] \\
Encode $[z^m_{t-1,k}, h^m_{t^*,k}, h^m_{t,k^*}, (\bar{a}_{t-1})]$ into $e^{m,\mathrm{prior}}_{t,k}$ & $\MLP^{\mathrm{prior}}_{\mathrm{mask}}$ & [128, 128] \\

Autoregressive posterior update of $h^m_{t,k^*\rightarrow k^*+1}$ from $[e^{m,\mathrm{post}}_{t,k}, z^m_{t,k}]$ & $\LSTM^{m,\mathrm{post}}$ & LSTM(96, 128) \\
Autoregressive prior update of $h^m_{t,k^*\rightarrow k^*+1}$ from $[h^m_{t^*,k}, z^m_{t,k}]$ & $\LSTM^{m,\mathrm{prior}}$ & LSTM(160, 128) \\

Infer $\bar{z}^m_{t,k}$ from $h^{m,\mathrm{post}}_{t,k^*}$ & $\MLP^{\mathrm{post}}_{q_\phi(z^m_{t,k}|\cdot)}$ & [128, 128] \\

Generate $\bar{z}^m_{t,k}$ from $[h^{m,\mathrm{prior}}_{t,k}, h^{m,\mathrm{prior}}_{t,k}]$ & $\MLP^{\mathrm{prior}}_{p_\theta(z^m_{t,k}|\cdot)}$ & [128, 128] \\
Residual update of $z^m_{t,k}$ from $[\bar{z}^m_{t,k}, z^m_{t-1,k}]$
& $\MLP^{\mathrm{mask}}_{\mathrm{res}}$ & [128, 128, 128]\\
Condition $\bar{a}^{t-1}$ on $z^m_{t,k}$ for decoding & $\MLP^{m,\mathrm{cond}}_{\mathrm{dec}}$ & [128, 128]\\
Decode $z^m_{t,k}$ into $\pi_{t,k}$ & $\CNN^\mathrm{mask}_{\mathrm{dec}}$ &  Refer to Table~\ref{tab:mask_decoder}.\\
Encode $\pi_{t,k}$ and $o_t$ into $e^{c,\mathrm{post}}_{t,k}$ & $\CNN^{\mathrm{comp}}_{\mathrm{enc}}$ & Refer to Table~\ref{tab:comp_encoder}.\\
Condition $h^c_{t,k}$ on $e^{c,\mathrm{post}}_{t,k}$ & $\MLP^{c,\mathrm{cond}}_{\mathrm{enc}}$ & [128, 128]\\
Condition $\bar{a}_{t-1}$ on $e^{c,\mathrm{post}}_{t,k}$ and infer $\bar{z}^c_{t,k}$ & $\MLP^{c,\mathrm{cond}}_{\mathrm{enc}}$ & [128, 128]\\
Generate $\bar{z}^c_{t,k}$ from $[z^m_{t,k}, z^c_{t-1,k}, h^c_{t,k}]$ & $\MLP^{c,\mathrm{prior}}_{\mathrm{comp}}$ & [128, ELU \cite{clevert2015fast}, 128, ELU, 128]\\
Residual update of $z^c_{t,k}$ from $[\bar{z}^c_{t,k}, z^c_{t-1,k}]$
& $\MLP^{\mathrm{comp}}_{\mathrm{res}}$ & [128, 128, 128]\\
Condition $\bar{a}^{t-1}$ on $z^c_{t,k}$ for decoding & $\MLP^{c,\mathrm{cond}}_{\mathrm{dec}}$ & [128, 128]\\
Decode $z^c_{t,k}$ into $o_{t,k}$ & $\CNN^\mathrm{comp}_{\mathrm{dec}}$ &  Refer to Table~\ref{tab:comp_decoder}.\\
Encode $\mu^{\mathrm{mix}}_t$ into $e^{\mathrm{mix}}_t$ & $\CNN^\mathrm{post}_{\mathrm{enc}}$ & Refer to Table~\ref{tab:post_encoder}.\\
Temporal update of $h^m_{t^*\rightarrow t^*+1,k}$ from $[z^m_{t,k}\sim q_\phi, e^{\mathrm{mix}}_t]$ & $\LSTM^{m,\mathrm{post}}_k$ in Eq.~(\ref{eq:separate_rnn}) & LSTM(160, 128) \\
Temporal update of $h^m_{t^*\rightarrow t^*+1,k}$ from $[z^m_{t,k}\sim p_\theta, e^{\mathrm{mix}}_t]$ & $\LSTM^{m,\mathrm{prior}}_k$ in Eq.~(\ref{eq:separate_rnn}) & LSTM(160, 128) \\
Temporal update of $h^c_{t^*\rightarrow t^*+1,k}$ from $[z^c_{t,k}\sim q_\phi, e^{\mathrm{mix}}_t]$ & $\LSTM^{c,\mathrm{post}}_k$ in Eq.~(\ref{eq:separate_rnn}) & LSTM(192, 128) \\
Temporal update of $h^c_{t^*\rightarrow t^*+1,k}$ from $[z^c_{t,k}\sim p_\theta, e^{\mathrm{mix}}_t]$ & $\LSTM^{c,\mathrm{prior}}_k$ in Eq.~(\ref{eq:separate_rnn}) & LSTM(192, 128) \\

\toprule
\textbf{Keypoint Module} \\
Condition $\bar{a}_{t-1}$ on $h^k_{t}$ & $\mathrm{Layer1}^{\mathrm{act}}_{\mathrm{cond}}$ & $512 + 64 \rightarrow 128$ \\
\midrule
\textbf{Interaction} \\
Extract global agent feature $e^t_{t,i}$ from $u^r_t = (z^r_t, h^r_t)$  & $f^{t,\mathrm{glob}}$ in Eq.~\ref{eq:temporal_interaction_global} & [128]\\
Integrate $e^t_{t,i}$ and $e^t_{t,i}$ as an ambient interaction of object $i$ & $\MLP^{\mathrm{amb}}$ & [128, 128, 32]\\
\bottomrule
\end{tabular}
\end{center}
\vskip -0.1in
}

%% file: tab/networks/generic/01_obs_encoder.tex
\begin{tabular}{llll}
\toprule
Layer           & Size/Ch. & Stride & Norm./Act.   \\ 
\midrule
Input           & 3       &        &              \\
Conv $7\times7$ & 64      & 2      & GN(4)/CELU \\
Conv $3\times3$ & 128      & 2      & GN(8)/CELU \\
Conv $3\times3$ & 256      & 2      & GN(16)/CELU \\
Conv $3\times3$ & 512      & 2      & GN(32)/CELU \\
Flatten & & \\
Linear & 128 &\\
\bottomrule
\end{tabular}
\vspace{-0.3cm}

%% file: tab/networks/mixture/mask/01_mask_decoder.tex
\begin{tabular}{llll}
\toprule
Layer           & Size/Ch. & Stride & Norm./Act.   \\ 
\midrule
Input           & 32 (1d)      &        &              \\
Conv $1\times1$ & 256      & 1      & GN(16)/CELU \\
SubConv $4\times4$ & 256   & 1      & GN(16)/CELU \\
Conv $3\times3$ & 256      & 1      & GN(16)/CELU \\
SubConv $2\times2$ & 128   & 1      & GN(16)/CELU \\
Conv $3\times3$ & 128      & 1      & GN(16)/CELU \\
SubConv $4\times4$ & 64    & 1      & GN(8)/CELU \\
Conv $3\times3$ & 64       & 1      & GN(8)/CELU \\
SubConv $4\times4$ & 16    & 1      & GN(4)/CELU \\
Conv $3\times3$ & 16       & 1      & GN(4)/CELU \\
Conv $3\times3$ & 1        & 1      &            \\
\bottomrule
\end{tabular}
\vspace{-0.3cm}

%% file: tab/networks/mixture/comp/01_comp_encoder.tex
\begin{tabular}{llll}
\toprule
Layer           & Size/Ch. & Stride & Norm./Act.   \\ 
\midrule
Input           & 3+1 (obs. + mask)      &        &              \\
Conv $3\times3$ & 32      & 2      & BN(32)/ELU \\
Conv $3\times3$ & 32      & 2      & BN(32)/ELU \\
Conv $3\times3$ & 64      & 2      & BN(64)/ELU \\
Flatten & & \\
Linear & 128 &\\
\bottomrule
\end{tabular}
\vspace{-0.3cm}

%% file: tab/networks/mixture/comp/02_comp_decoder.tex
\begin{tabular}{llll}
\toprule
Layer           & Size/Ch. & Stride & Norm./Act.   \\ 
\midrule
Input           & 64 (1d)      &        &              \\
Spatial Broadcast & 64 + 2 (1d)      &        &              \\
Conv $3\times3$ & 32      & 1      & BN(32)/ELU \\
Conv $3\times3$ & 32      & 1      & BN(32)/ELU \\
Conv $3\times3$ & 32      & 1      & BN(32)/ELU \\
Conv $3\times3$ & 3      & 1      &             \\
\bottomrule
\end{tabular}
\vspace{-0.3cm}

%% file: tab/hyperparam/hyper.tex
\begin{center}
\begin{small}
\begin{tabular}{lll}
\toprule
 Functionality & Notation & Value \\
\midrule
Image size & $(H, W)$ & (64, 64)\\
Sample length & $T$ & $[5, 7, \cdots, 25, 27, 30]$ \\
Sample milestones & & $[20k, 30k, \cdots, 100k, 110k]$\\
Learning rate decay & & 0.8 \\
Lr decay milestones & & $[100K, 150K]$ \\
Num. of keypoints  & $N$ & 32 \\
Dim. of $h^k_t$ &  & 512 \\
Dim. of $z^k_{t,n}$ &  & 16 \\
Num. of samples for best belief &  & 50 \\
Keypoint sep. loss scale & $\lambda_{\mathrm{sep}}$ & 0.02 \\
Keypoint sparse loss scale & $\lambda_{\mathrm{sparse}}$ & 0.002 \\
Keypoint KL loss scale &  & 0.001 \\
Heatmap regularization scale & $\lambda$ & $0.01$ \\
Dim. of enhanced action & $\bar{a}_t$ & 32 \\
Dim. of $z^m_{t,k}$ &  & 32 \\
Dim. of $h^m_{t,k}$ &  & 128 \\
Dim. of $z^c_{t,k}$ &  & 64 \\
Dim. of $h^c_{t,k}$ &  & 128 \\
$o_t$ standard deviation & $\sigma$ & 0.1 \\
\bottomrule
\end{tabular}
\end{small}
\end{center}
\vspace{-0.6cm}

%% file: tab/hyperparam/dataset/01_rob_obj_hyper.tex
\begin{center}
\begin{small}
\begin{tabular}{lll}
\toprule
 Functionality & Notation & Value \\
\midrule
Optimizer &  & Adam \cite{kingma2014adam} \\
Learning rate start & & $3 \times 10^{-4}$ \\
Batch size & & $4$ \\
Num. of mixture modes & $K$ & $3$ \\
Keypoint only training steps & & $0 \sim 80k$ \\
Mixture only training steps & & $80k \sim 110k$ \\
Mixture-keypoint joint steps & & $80k \sim 300k$ \\
Fix alpha $\alpha$ steps &  & $110k \sim 120k$ \\
Fixed alpha value & $\alpha$ & 0.45 \\
Mixture standard deviation & $ \sigma^{\mathrm{mix}} $ & $0.1$ \\
Discovery grid divisions & $ G $ & $4$ \\ 
Max num. of object discovery & & $7$ \\
Raw action dimension & $a_t$ & $7$ \\
Residual regularization & $\lambda$ & $1.0$ \\
Residual scale & $\Delta$ & $2.0$ \\
\bottomrule
\end{tabular}
\end{small}
\end{center}
\vspace{-0.3cm}

%% file: tab/hyperparam/dataset/02_panda_hyper.tex
\begin{center}
\begin{small}
\begin{tabular}{lll}
\toprule
 Functionality & Notation & Value \\
\midrule
Optimizer &  & Adam \\
Learning rate start & & $4 \times 10^{-4}$ \\
Batch size & & $4$ \\
Num. of mixture modes & $K$ & $3$ \\
Keypoint only training steps & & $0 \sim 80k$ \\
Mixture only training steps & & $80k \sim 120k$ \\
Mixture-keypoint joint steps & & $80k \sim 1000k$ \\
Fix alpha $\alpha$ steps &  & $120k \sim 140k$ \\
Fixed alpha value & $\alpha$ & 0.4 \\
Mixture standard deviation & $ \sigma^{\mathrm{mix}} $ & $0.1$ \\
Discovery grid divisions & $ G $ & $4$ \\ 
Max num. of object discovery & & $7$ \\
Raw action dimension & $a_t$ & $7$ \\
Residual regularization & $\lambda$ & $0.01$ \\
Residual scale & $\Delta$ & $2.0$ \\
\bottomrule
\end{tabular}
\end{small}
\end{center}
\vspace{-0.6cm}

%% file: tab/hyperparam/dataset/03_sawyer_hyper.tex
\begin{center}
\begin{small}
\begin{tabular}{lll}
\toprule
 Functionality & Notation & Value \\
\midrule
Optimizer &  & Adam \\
Learning rate start & & $4 \times 10^{-4}$ \\
Batch size & & $4$ \\
Num. of mixture modes & $K$ & $3$ \\
Keypoint only training steps & & $0 \sim 80k$ \\
Mixture only training steps & & $80k \sim 100k$ \\
Mixture-keypoint joint steps & & $80k \sim 900k$ \\
Fix alpha $\alpha$ steps &  & $100k \sim 110k$ \\
Fixed alpha value & $\alpha$ & 0.4 \\
Mixture standard deviation & $ \sigma^{\mathrm{mix}} $ & $0.5$ \\
Discovery grid divisions & $ G $ & $4$ \\ 
Max num. of object discovery & & $7$ \\
Raw action dimension & $a_t$ & $7$ \\
Residual regularization & $\lambda$ & $0.01$ \\
Residual scale & $\Delta$ & $1.0$ \\
\bottomrule
\end{tabular}
\end{small}
\end{center}
\vspace{-0.6cm}

%% file: tab/hyperparam/dataset/04_bair_hyper.tex
\begin{center}
\begin{small}
\begin{tabular}{lll}
\toprule
 Functionality & Notation & Value \\
\midrule
Optimizer &  & Adam \\
Learning rate start & & $4 \times 10^{-4}$ \\
Batch size & & $4$ \\
Num. of mixture modes & $K$ & $4$ \\
Keypoint only training steps & & $0 \sim 80k$ \\
Mixture only training steps & & $80k \sim 110k$ \\
Mixture-keypoint joint steps & & $80k \sim 160k$ \\
Fix alpha $\alpha$ steps &  & $100k \sim 110k$ \\
Fixed alpha value & $\alpha$ & 0.4 \\
Mixture standard deviation & $ \sigma^{\mathrm{mix}} $ & $0.5$ \\
Discovery grid divisions & $ G $ & $8$ \\ 
Max num. of object discovery & & $12$ \\
Raw action dimension & $a_t$ & $3$ \\
Residual regularization & $\lambda$ & $1.0$ \\
Residual scale & $\Delta$ & $2.0$ \\
\bottomrule
\end{tabular}
\end{small}
\end{center}
\vspace{-0.8cm}

%% file: 05_additional_samples.tex
\newpage
\onecolumn
\section{Additional Samples}
In this section, we present additional qualitative and quantitative results of the experiments conducted in the Sec.~\ref{sec:Experiments} of the main paper.
\label{sec:additional_samples}
\subsection{Spatial Decomposition}
As we have only shown the comparison of spatial decomposition for all methods on the ROLL dataset in the Sec.~\ref{subsec:spatial_decomp} of the main paper, we report additional results for BAIR (Fig.~\ref{fig:decomp_bair}), PUSH1 (Fig.~\ref{fig:decomp_panda}), and PUSH2 (Fig.~\ref{fig:decomp_sawyer}) datasets. Only GATSBI can decompose a scene into the agent, background entities, and objects.
\input{fig/sample/00_spatial_decomp/00_spatial_decomp_bair}

\input{fig/sample/00_spatial_decomp/01_spatial_decomp_panda}

\input{fig/sample/00_spatial_decomp/02_spatial_decomp_sawyer}

\newpage
\subsection{Temporal Prediction}
On top of the temporal prediction results of PUSH1 and BAIR datasets introduced in Sec.~\ref{subsec:spatio_temp} of the main paper, we show additional results for ROLL (Fig.~\ref{fig:temp_rob_obj}) and PUSH2 (Fig.~\ref{fig:temp_sawyer}) datasets.
\input{fig/sample/01_temporal/00_temporal_rob_obj}
\input{fig/sample/01_temporal/01_temporal_sawyer}

\newpage
\subsection{Spatio-temporal Prediction}
GATSBI can robustly propagate the spatially decomposed scene entities over the time steps. The temporally consistent prediction for each entity is essential for the robustness of the method. Therefore, we present the spatio-temporal prediction results of GATSBI on the four robotics datasets 
(Fig.~\ref{fig:spatiotemp_bair} to Fig.~\ref{fig:spatiotemp_rob_obj}).
\input{fig/sample/02_spatio_temp/01_spatiotemp_bair}
\input{fig/sample/02_spatio_temp/02_spatiotemp_panda}
\input{fig/sample/02_spatio_temp/03_spatiotemp_sawyer}
\input{fig/sample/02_spatio_temp/04_spatiotemp_rob_obj}

\newpage
\subsection{Physically Plausible Samples}
In the case of the prediction of the agent-object interaction does not coincide with the ground truth, GATSBI still generates the physically plausible predictions of the complex interactions. As shown in Fig.~\ref{fig:physics_panda} and Fig.~\ref{fig:physics_sawyer}, all the three different runs for the same scenarios generate reasonable physical interactions among the scene entities.
\input{fig/sample/03_physics_plausible/01_physics_panda}
\input{fig/sample/03_physics_plausible/02_physics_sawyer}

\newpage
\subsection{Quantitative Evaluations}
In addition to the quantitative results discussed in Sec.~\ref{sec:Experiments}, we report various numerical metrics for evaluating the video prediction qualities of the methods we test on. GATSBI outperforms all the other methods in various qualitative metrics (Fig.~\ref{fig:rob_obj_metrics} to Fig.~\ref{fig:bair_metrics}).
\input{fig/metrics/01_rob_obj_metrics}

\input{fig/metrics/02_panda_metrics}

\input{fig/metrics/03_sawyer_metrics}

\input{fig/metrics/04_bair_metrics}

\newpage
\subsection{Additional Ablation Study}
\paragraph{Separation of RNN between Prior and Posterior}
We test how the separation of the RNN of posterior and prior distribution improves the performance of predicting the trajectory of the agent (Fig.~\ref{fig:rnn_split_rob_obj} and Fig.~\ref{fig:rnn_split_panda}). The results show that training individual RNN for each posterior and prior distribution yields better results for the accuracy in predicting the agent trajectory.
\input{fig/ablation/rnn_split/01_rnn_split_rob_obj}
\input{fig/ablation/rnn_split/02_rnn_split_panda}
\newpage
\paragraph{Joint Action Conditioning between Mixture and Keypoint Modules}
We also test the efficacy of joint training of the agent latent variable and the keypoint latent variable.  Fig.~\ref{fig:kypt_cond_rob_obj} to  Fig.~\ref{fig:kypt_cond_bair} demonstrates that jointly training the two modules improves the prediction accuracy of the agent trajectory. 
\input{fig/ablation/kypt_cond/01_kypt_cond_rob_obj}
\input{fig/ablation/kypt_cond/02_kypt_cond_panda}
\input{fig/ablation/kypt_cond/03_kypt_cond_sawyer}
\input{fig/ablation/kypt_cond/04_kypt_cond_bair}

\newpage
\subsection{Agent-free Interactions}
\input{fig/sample/04_balls_inter/01_balls_inter}
We finally report the samples from the agent-free interaction scenario discussed in Sec.~\ref{parag:balls_inter} of the main paper. Fig.~\ref{fig:balls_inter} shows a physically plausible prediction of the interaction among objects.
\newpage

%% file: fig/sample/00_spatial_decomp/00_spatial_decomp_bair.tex
\begin{figure*}[!ht]
    \centering
    \includegraphics[scale=1.0]{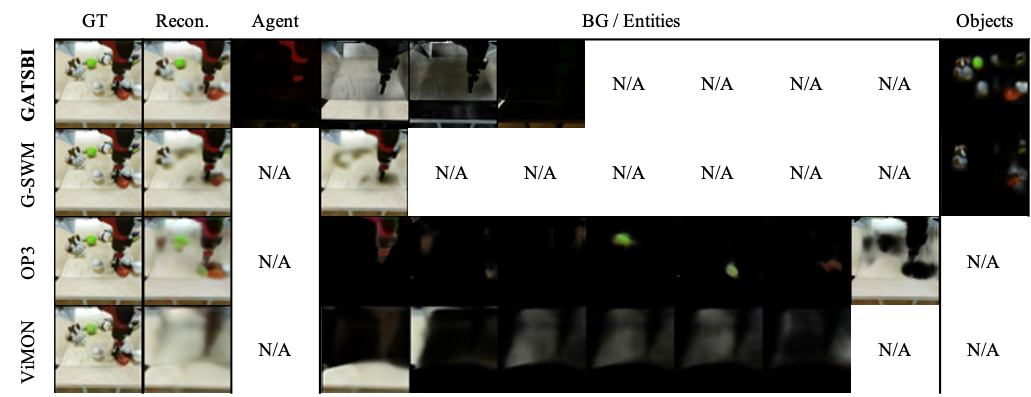}
    \caption{Spatial decomposition results on BAIR dataset.}
    \label{fig:decomp_bair}
\end{figure*}

%% file: fig/sample/00_spatial_decomp/01_spatial_decomp_panda.tex
\begin{figure*}[!ht]
    \centering
    \includegraphics[scale=1.0]{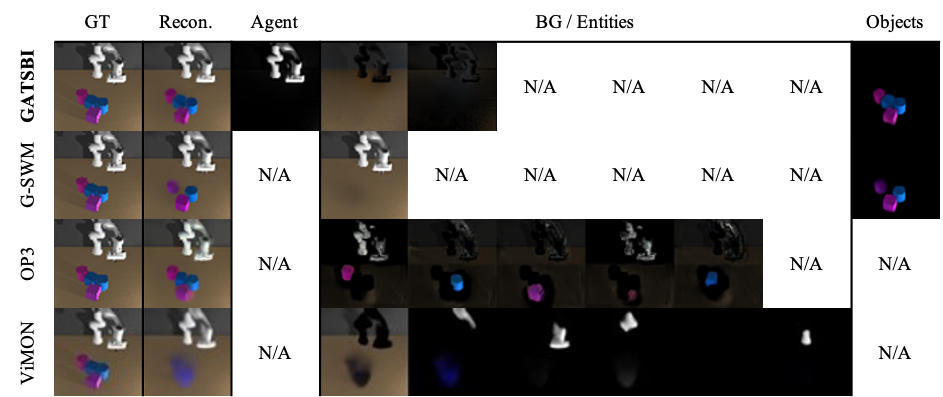}
    \caption{Spatial decomposition results on PUSH1 dataset.}
    \label{fig:decomp_panda}
\end{figure*}

%% file: fig/sample/00_spatial_decomp/02_spatial_decomp_sawyer.tex
\begin{figure*}[!ht]
    \centering
    \includegraphics[scale=1.0]{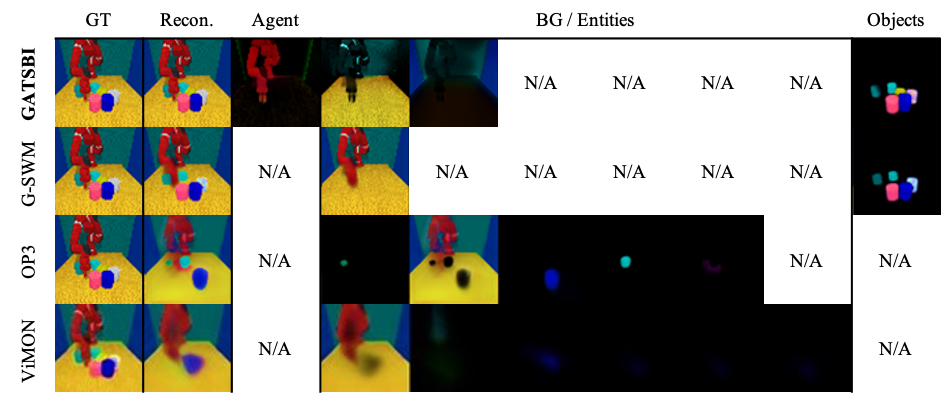}
    \caption{Spatial decomposition results on PUSH2 dataset.}
    \label{fig:decomp_sawyer}
\end{figure*}

%% file: fig/sample/01_temporal/00_temporal_rob_obj.tex
\begin{figure*}[!ht]
    \vspace{-0.5cm}
    \centering
    \includegraphics[scale=1.0]{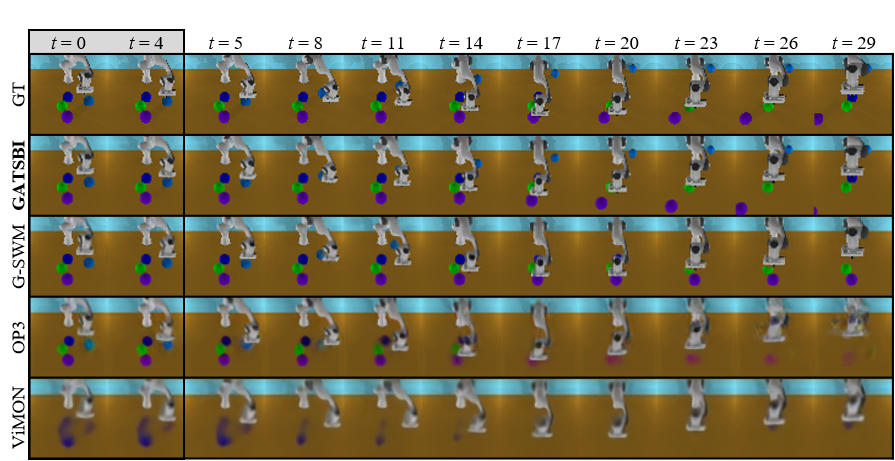}
    \caption{Temporal prediction results on ROLL dataset.}
    \vspace{-0.3cm}
    \label{fig:temp_rob_obj}
\end{figure*}

%% file: fig/sample/01_temporal/01_temporal_sawyer.tex
\begin{figure*}[!ht]
    \vspace{-0.5cm}
    \centering
    \includegraphics[scale=1.0]{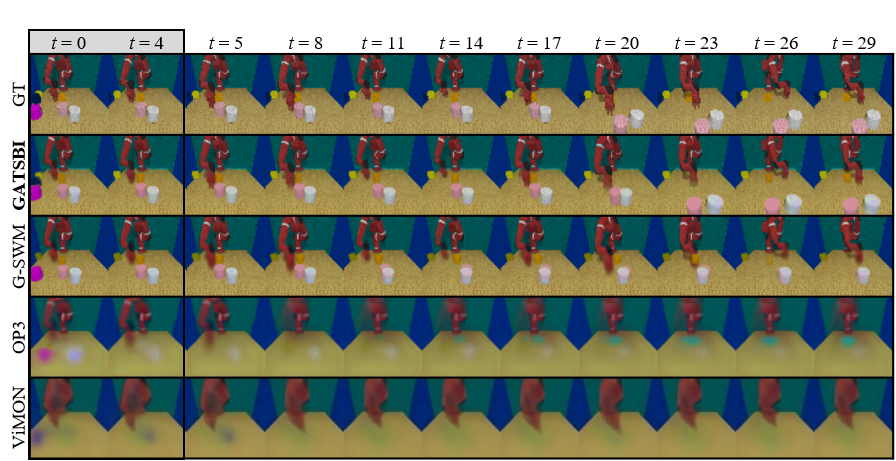}
    \caption{Temporal prediction results on PUSH2 dataset.}
    \vspace{-0.3cm}
    \label{fig:temp_sawyer}
\end{figure*}

%% file: fig/sample/02_spatio_temp/01_spatiotemp_bair.tex
\begin{figure}[!ht]
    \vspace{-0.5cm}
    \centering
    \begin{subfigure}
        \centering
        \includegraphics[scale=1.0]{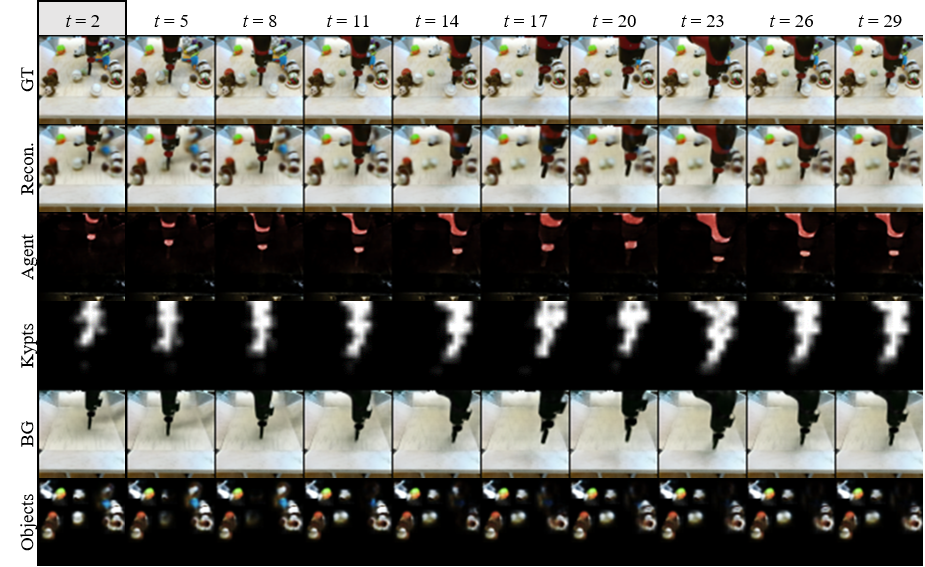}
        Spatio-temporal decomposition on episode A.
    \end{subfigure}
    \vfill
    \vspace{0.1cm}
    \begin{subfigure}
        \centering
        \includegraphics[scale=1.0]{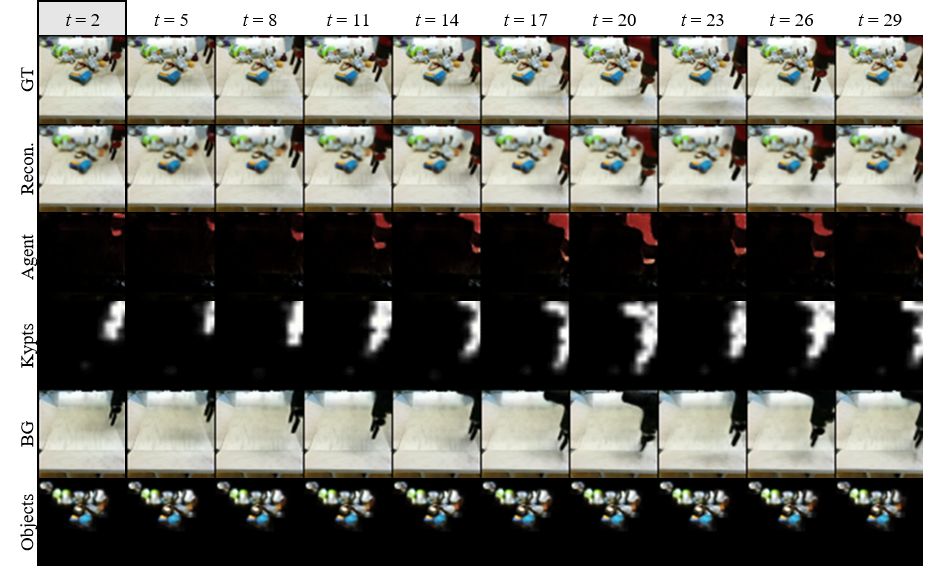}
        Spatio-temporal decomposition on episode B.
    \end{subfigure}
    \vspace{0.3cm}
    \caption{Spatio-temporal prediction for running GATSBI on two different episodes from BAIR dataset. GATSBI consistently propagates the spatially decomposed scene entities along the time horizon.}
    \label{fig:spatiotemp_bair}

\end{figure}

%% file: fig/sample/02_spatio_temp/02_spatiotemp_panda.tex
\begin{figure}[!ht]
    \vspace{-0.5cm}
    \centering
    \begin{subfigure}
        \centering
        \includegraphics[scale=1.0]{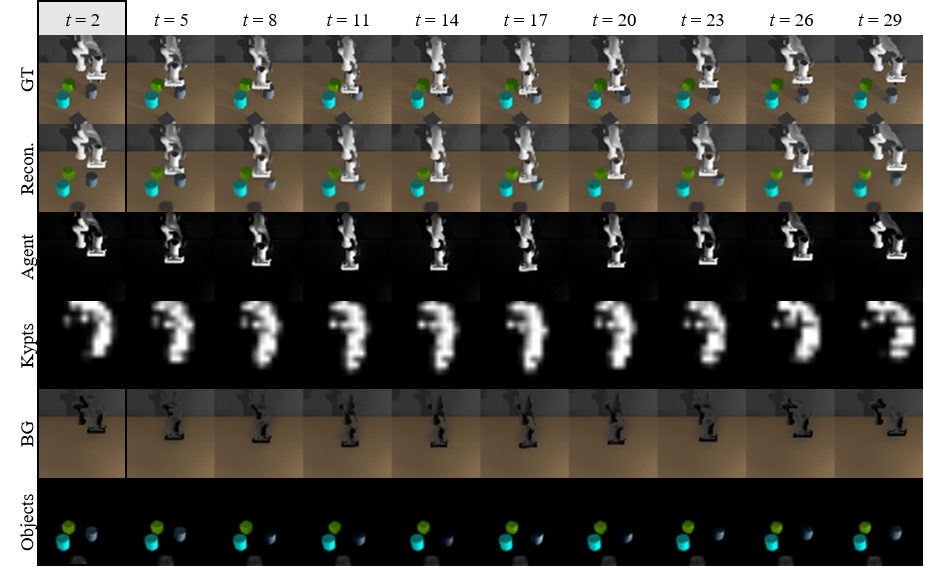}
        Spatio-temporal decomposition on episode A.
    \end{subfigure}
    \vfill
    \vspace{0.1cm}
    \begin{subfigure}
        \centering
        \includegraphics[scale=1.0]{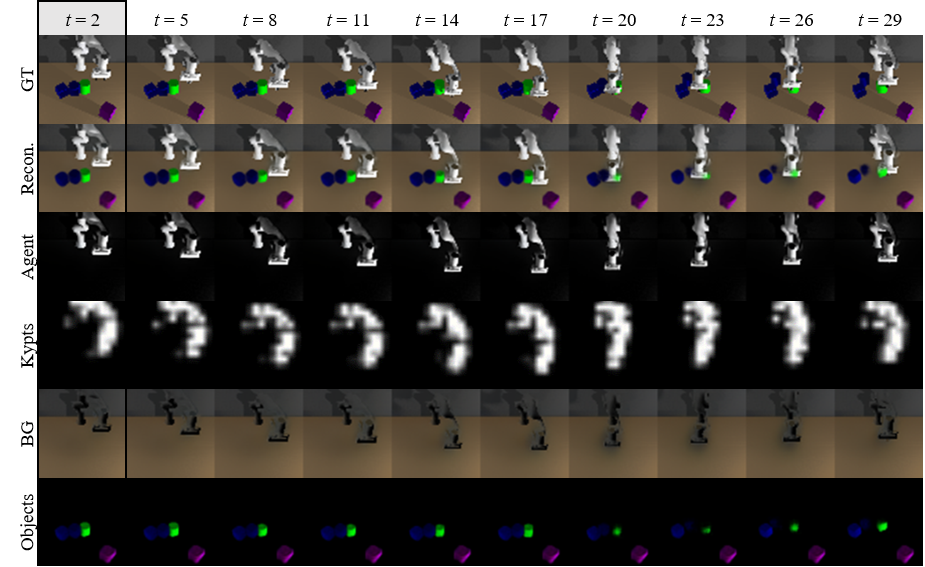}
        Spatio-temporal decomposition on episode B.
    \end{subfigure}
    \vspace{0.3cm}
    \caption{Spatio-temporal prediction for running GATSBI on two different episodes from PUSH1 dataset. GATSBI consistently propagates the spatially decomposed scene entities along the time horizon.}
    \label{fig:spatiotemp_panda}

    \vspace{0.5cm}
\end{figure}

%% file: fig/sample/02_spatio_temp/03_spatiotemp_sawyer.tex
\begin{figure}[!ht]
    \vspace{-0.5cm}
    \centering
    \begin{subfigure}
        \centering
        \includegraphics[scale=1.0]{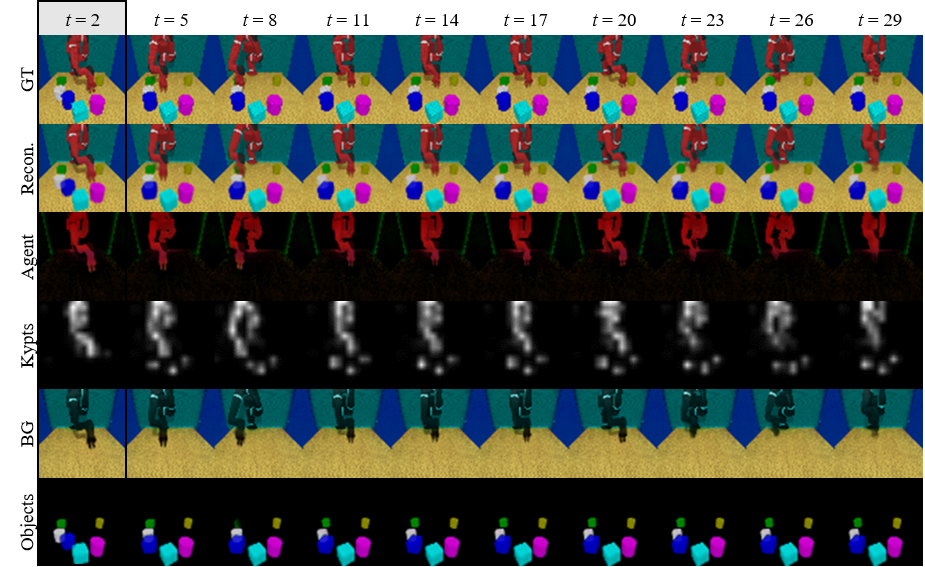}
        Spatio-temporal decomposition on episode A.
    \end{subfigure}
    \vfill
    \vspace{0.1cm}
    \begin{subfigure}
        \centering
        \includegraphics[scale=1.0]{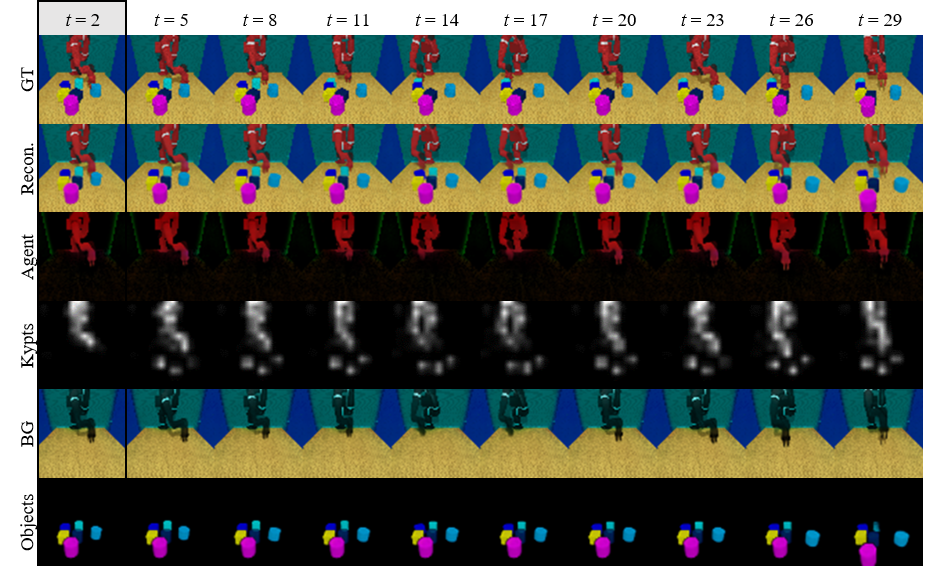}
        Spatio-temporal decomposition on episode B.
    \end{subfigure}
    
    \vspace{0.3cm}
    \caption{Spatio-temporal prediction for running GATSBI on two different episodes from PUSH2 dataset. GATSBI consistently propagates the spatially decomposed scene entities along the time horizon.}
    \label{fig:spatiotemp_sawyer}
    \vspace{0.5cm}
\end{figure}

%% file: fig/sample/02_spatio_temp/04_spatiotemp_rob_obj.tex
\begin{figure}[!ht]
    \vspace{-0.5cm}
    \centering
    \begin{subfigure}
        \centering
        \includegraphics[scale=1.0]{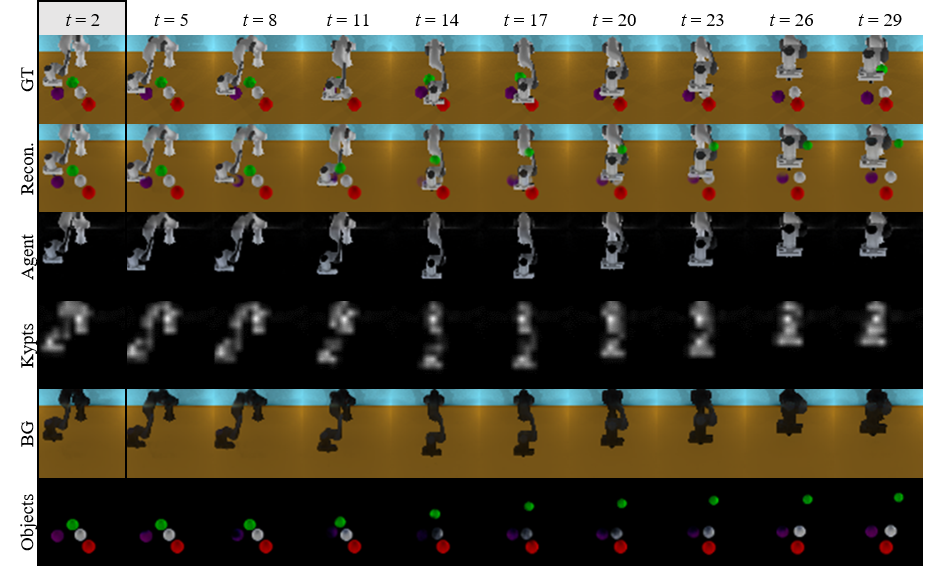}
        Spatio-temporal decomposition on episode A.
    \end{subfigure}
    \vfill
    \vspace{0.1cm}
    \begin{subfigure}
        \centering
        \includegraphics[scale=1.0]{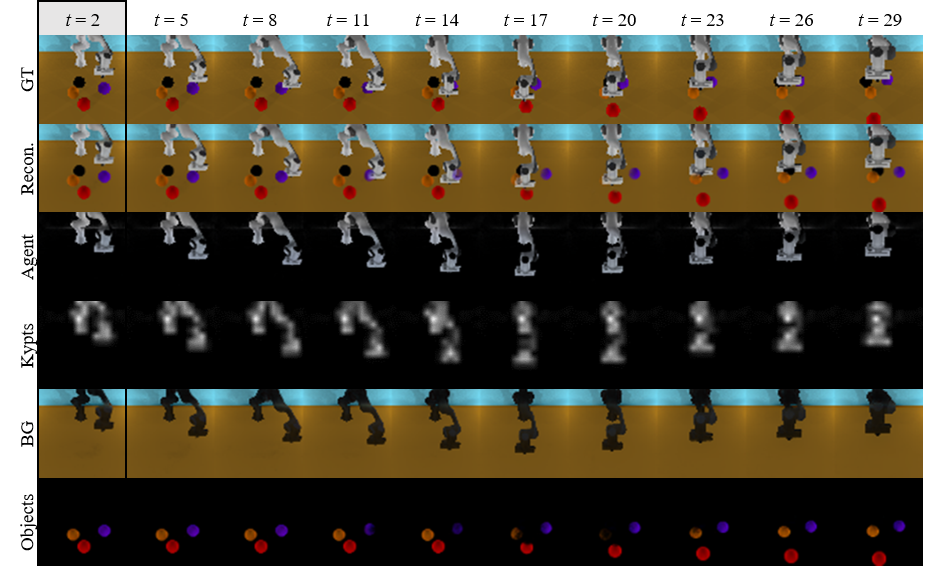}
        Spatio-temporal decomposition on episode B.
    \end{subfigure}
    \vspace{0.3cm}
    \caption{Spatio-temporal prediction for running GATSBI on two different episodes from ROLL dataset. GATSBI consistently propagates the spatially decomposed scene entities along the time horizon.}
    \label{fig:spatiotemp_rob_obj}
    \vspace{0.5cm}
\end{figure}

%% file: fig/sample/03_physics_plausible/01_physics_panda.tex
\begin{figure}[!ht]
    \vspace{0.5cm}
    \centering
    \begin{subfigure}
        \centering
        \includegraphics[scale=1.0]{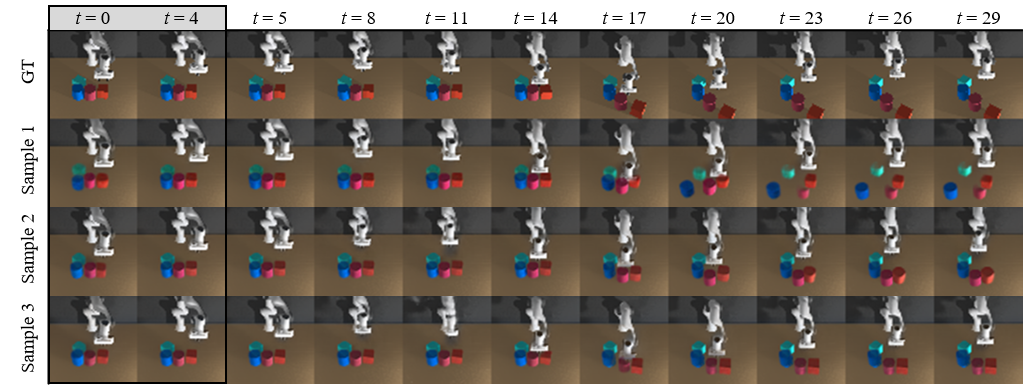}
        Three different runs for episode A.
    \end{subfigure}
    \vfill
    \vspace{0.1cm}
    \begin{subfigure}
        \centering
        \includegraphics[scale=1.0]{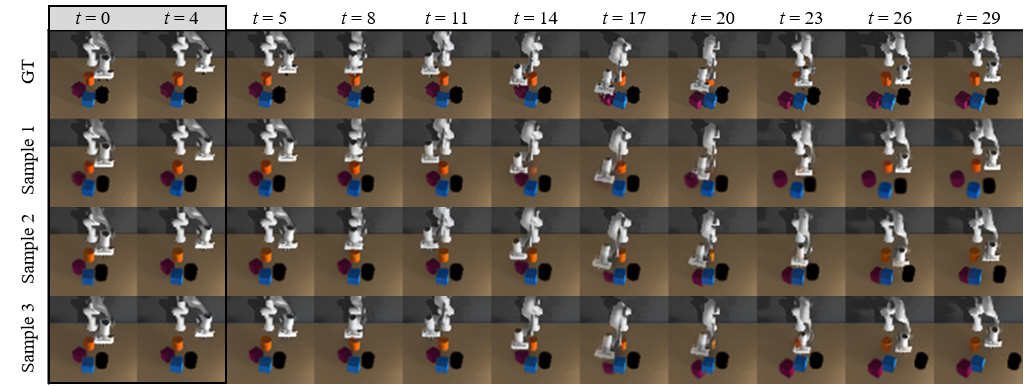}
        Three different runs for episode B.
    \end{subfigure}
    \vspace{0.3cm}
    \caption{Physically plausible samples for running GATSBI on two different episodes from PUSH1 dataset. For each run, GATSBI predicts frames that yields different agent-object interaction results with the ground truth.}
    \label{fig:physics_panda}
\end{figure}

%% file: fig/sample/03_physics_plausible/02_physics_sawyer.tex
\begin{figure}[!ht]
    \vspace{-0.5cm}
    \centering
    \begin{subfigure}
        \centering
        \includegraphics[scale=1.0]{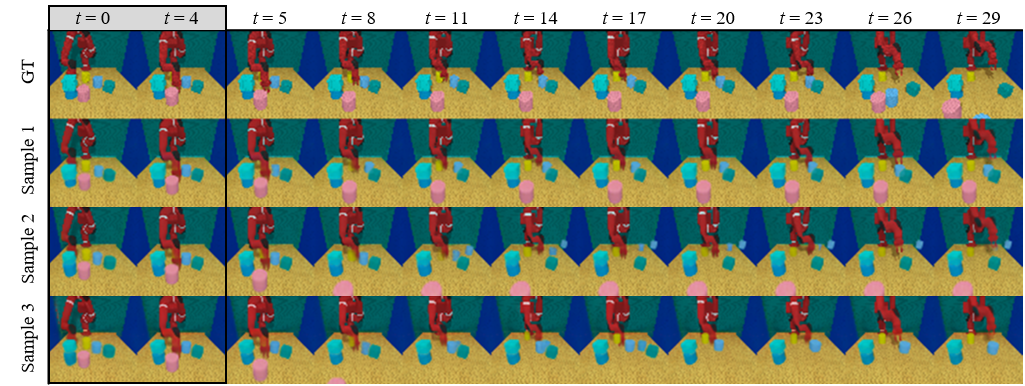}
        Three different runs for episode A.
    \end{subfigure}
    \vfill
    \vspace{0.1cm}
    \begin{subfigure}
        \centering
        \includegraphics[scale=1.0]{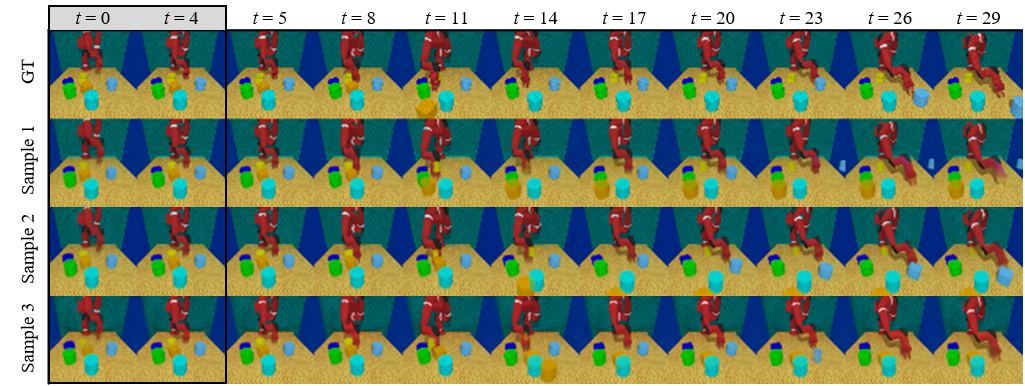}
        Three different runs for episode B.
    \end{subfigure}
    \begin{subfigure}
        \centering
        \includegraphics[scale=1.0]{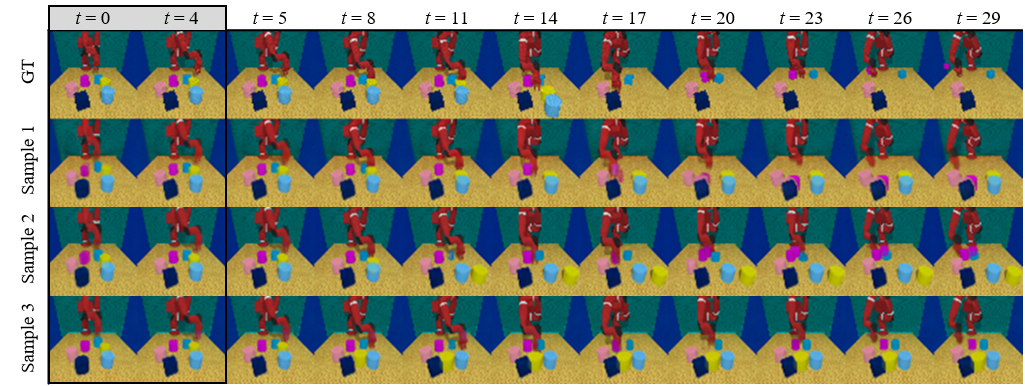}
        Three different runs for episode C.
    \end{subfigure}
    \vspace{0.3cm}
    \caption{Physically plausible samples for running GATSBI on two different episodes from PUSH2 dataset. For each run, GATSBI predicts frames that yields different agent-object interaction results with the ground truth.}
    \label{fig:physics_sawyer}
\end{figure}

%% file: fig/metrics/01_rob_obj_metrics.tex
\begin{figure*}[!ht]
    \centering
    \includegraphics[scale=0.45]{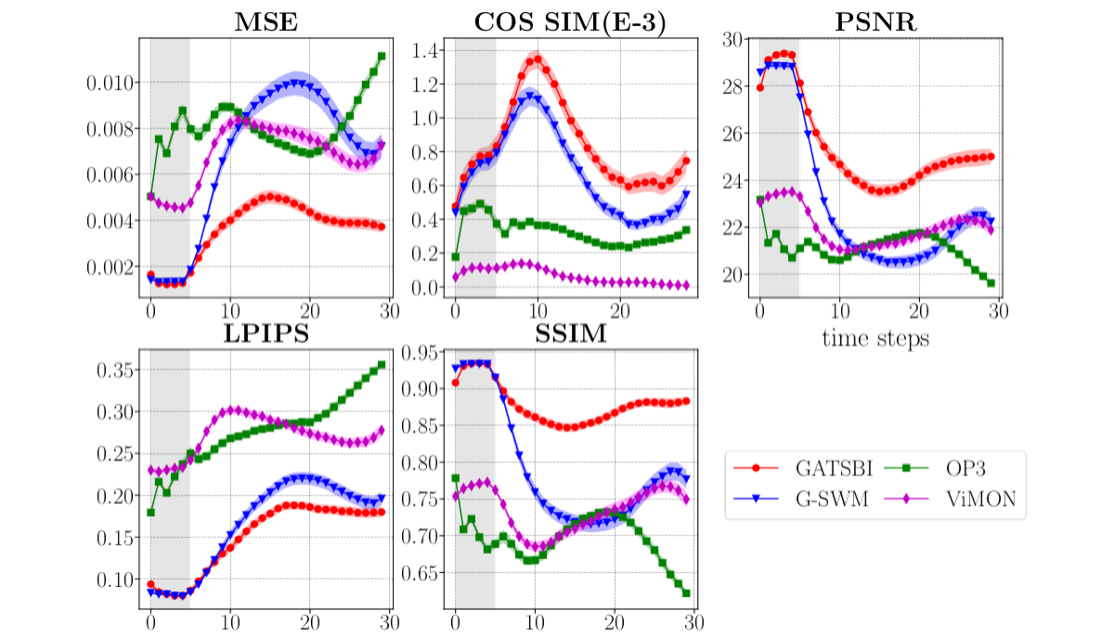}
    \caption{MSE and LPIPS (\emph{lower is better}), cosine similarity, PSNR, and LPIPS (\emph{higher is better}) along the time steps for ROLL dataset. Shaded area is the conditioning steps.}
    \label{fig:rob_obj_metrics}
\end{figure*}

%% file: fig/metrics/02_panda_metrics.tex
\begin{figure*}[!ht]
    \centering
    \includegraphics[scale=0.45]{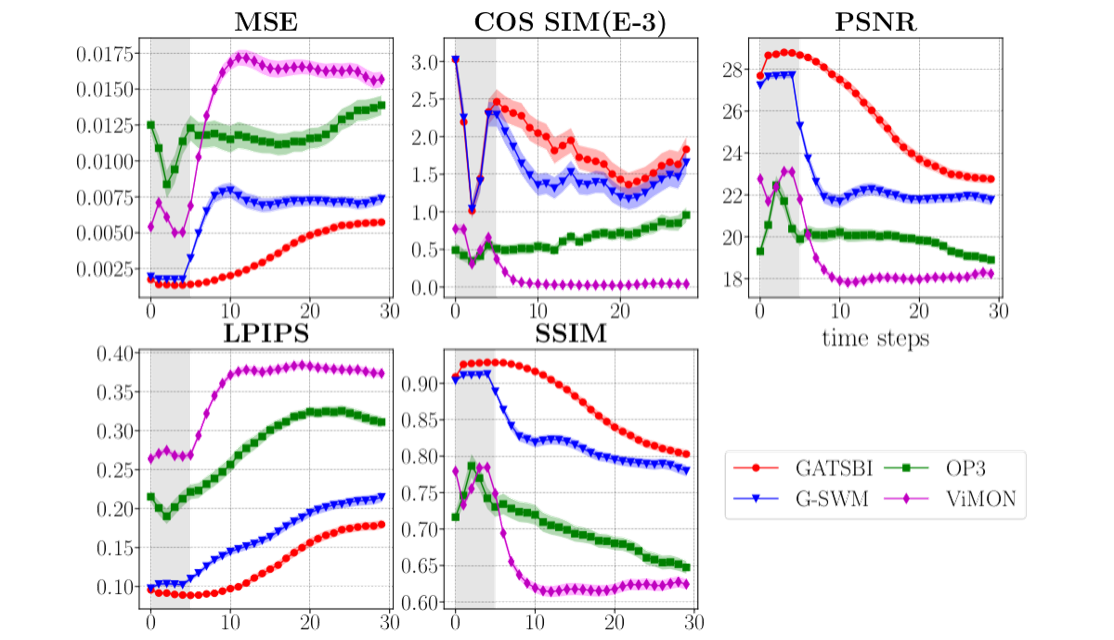}
    \caption{MSE and LPIPS (\emph{lower is better}), cosine similarity, PSNR, and LPIPS (\emph{higher is better}) aong the time steps for PUSH1 dataset. Shaded area is the conditioning steps.}
    \label{fig:panda_metrics}
\end{figure*}

%% file: fig/metrics/03_sawyer_metrics.tex
\begin{figure*}[!ht]
    \centering
    \includegraphics[scale=0.5]{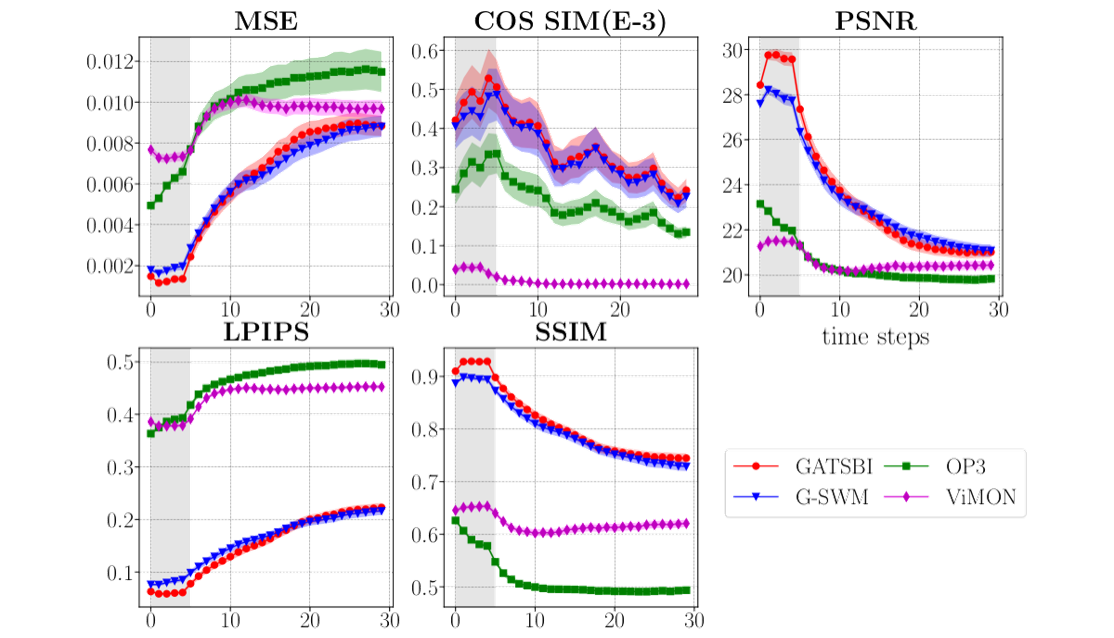}
    \caption{MSE and LPIPS (\emph{lower is better}), Cosine similarity, PSNR, and LPIPS (\emph{higher is better}) along the time steps for PUSH2 dataset. Shaded area is the conditioning steps.}
    \label{fig:sawyer_metrics}
\end{figure*}

%% file: fig/metrics/04_bair_metrics.tex
\begin{figure*}[!ht]
    \centering
    \includegraphics[scale=0.5]{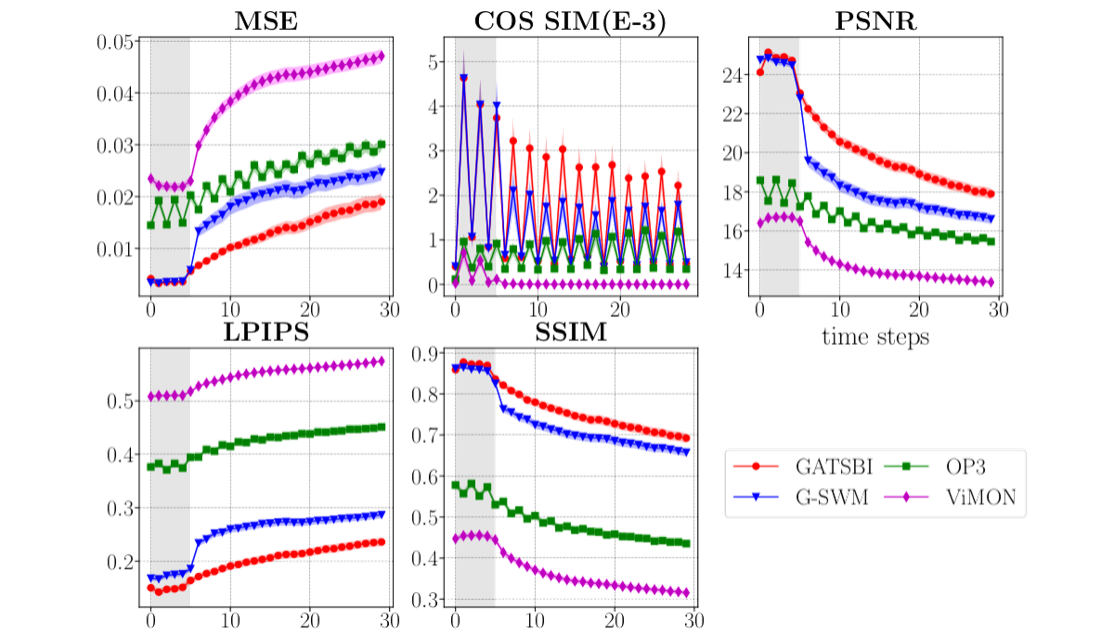}
    \caption{MSE and LPIPS (\emph{lower is better}), Cosine similarity, PSNR, and LPIPS (\emph{higher is better}) along the time steps for BAIR dataset. Shaded area is the conditioning steps.}
    \label{fig:bair_metrics}
\end{figure*}

%% file: fig/ablation/rnn_split/01_rnn_split_rob_obj.tex
\begin{figure}[!ht]
    \vspace{-0.5cm}
    \centering
    \begin{subfigure}
        \centering
        \includegraphics[scale=0.9]{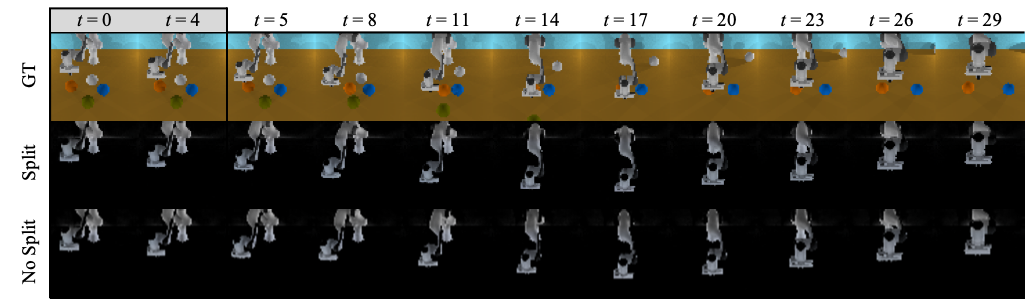}\\
        Agent trajectory comparison in ROLL dataset.
    \end{subfigure}
    \vfill
    \vspace{0.1cm}
    \begin{subfigure}
        \centering
        \includegraphics[scale=0.9]{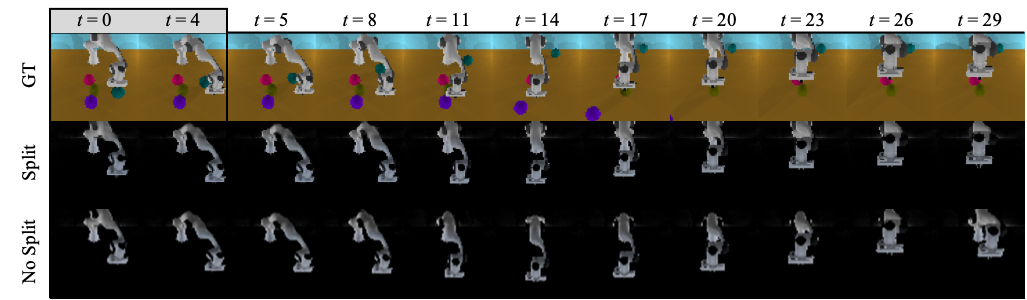}\\
        Agent trajectory comparison in ROLL dataset.
    \end{subfigure}
    \caption{Comparison between the trajectory of the agent of ROLL dataset, for each implementation of RNN in mixture module. From first to third row, groud truth observation, trajectory prediction with separated RNN, and integrated RNN are shown.}
    \label{fig:rnn_split_rob_obj}
\end{figure}

%% file: fig/ablation/rnn_split/02_rnn_split_panda.tex
\begin{figure}[!ht]
    \vspace{-0.5cm}
    \centering
    \begin{subfigure}
        \centering
        \includegraphics[scale=0.9]{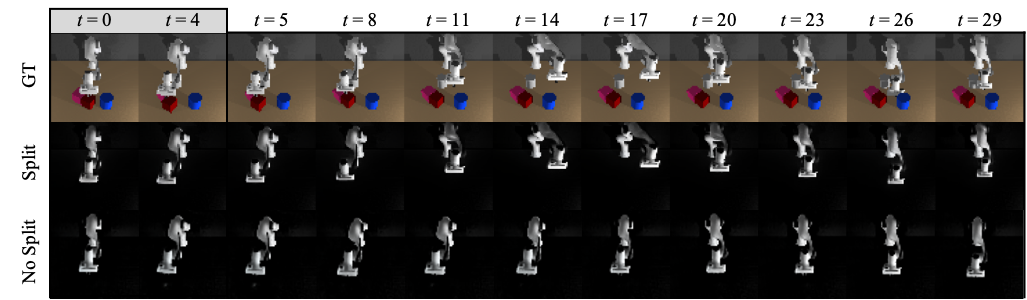}\\
        Agent trajectory comparison in PUSH1 dataset.
    \end{subfigure}
    \vfill
    \vspace{0.1cm}
    \begin{subfigure}
        \centering
        \includegraphics[scale=0.9]{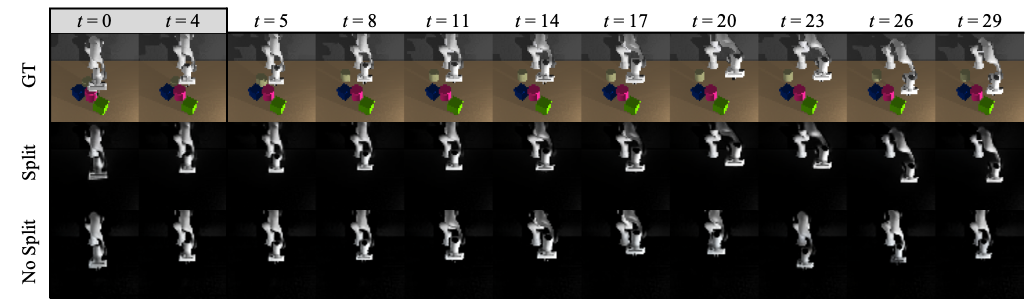}\\
        Agent trajectory comparison in PUSH1 dataset.
    \end{subfigure}
    \caption{Comparison between the trajectory of the agent of PUSH1 dataset, for each implementation of RNN in mixture module. From first to third row, groud truth observation, trajectory prediction with separated RNN, and integrated RNN are shown.}
    \label{fig:rnn_split_panda}
\end{figure}

%% file: fig/ablation/kypt_cond/01_kypt_cond_rob_obj.tex
\begin{figure}[!ht]    
\vspace{-0.3cm}
    \centering
    \includegraphics[scale=0.85]{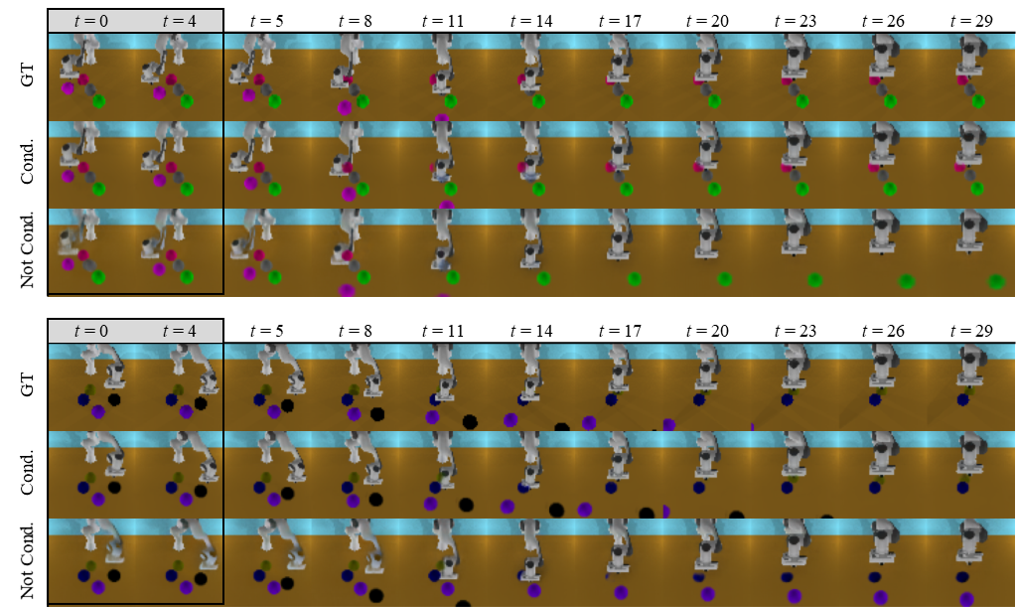}
    \vspace{0.3cm}
    \caption{Comparison between the trajectory of the agent of ROLL dataset, with and without latent dynamics sharing with keypoint module. From first to third row, groud truth observation, trajectory prediction with shared dynamics, and mask latent dynamics alone are shown.}
        \label{fig:kypt_cond_rob_obj}
    \vspace{-0.5cm}
\end{figure}

%% file: fig/ablation/kypt_cond/02_kypt_cond_panda.tex
\begin{figure}[!ht]
    \vspace{-0.3cm}
    \centering
    \includegraphics[scale=0.85]{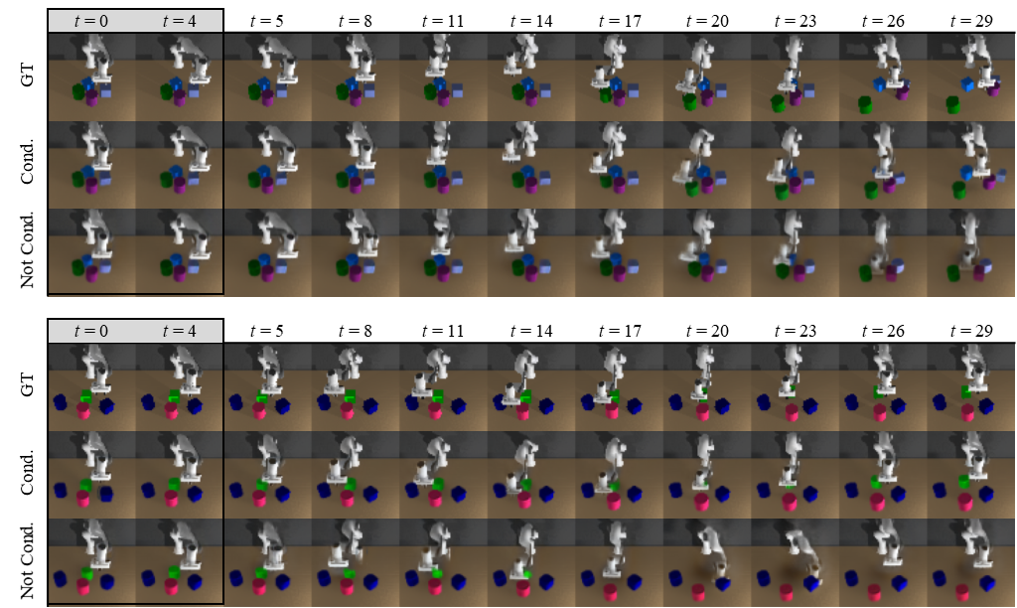}
    \vspace{0.3cm}
    \caption{Comparison between the trajectory of the agent of PUSH1 dataset, with and without latent dynamics sharing with keypoint module. From first to third row, groud truth observation, trajectory prediction with shared dynamics, and mask latent dynamics alone are shown.}
        \label{fig:kypt_cond_panda}
    \vspace{-0.5cm}
\end{figure}

%% file: fig/ablation/kypt_cond/03_kypt_cond_sawyer.tex
\begin{figure}[!ht]
    \vspace{-0.5cm}
    \centering
    \includegraphics[scale=0.85]{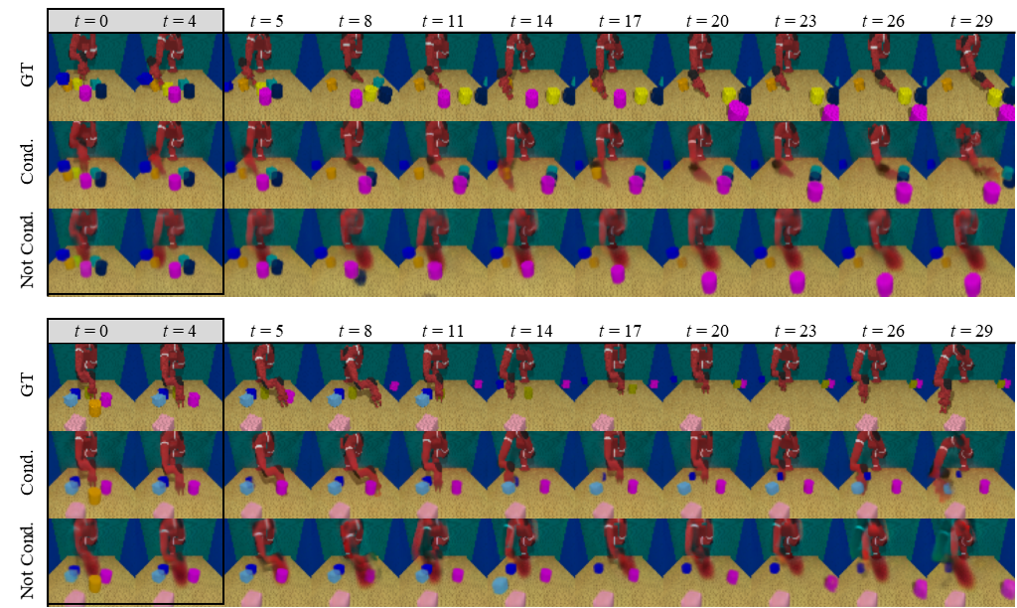}
    \vspace{0.3cm}
    \caption{Comparison between the trajectory of the agent of PUSH2 dataset, with and without latent dynamics sharing with keypoint module. From first to third row, groud truth observation, trajectory prediction with shared dynamics, and mask latent dynamics alone are shown.}
        \label{fig:kypt_cond_sawyer}
\end{figure}

%% file: fig/ablation/kypt_cond/04_kypt_cond_bair.tex
\begin{figure}[!ht]
    \centering
    \includegraphics[scale=0.85]{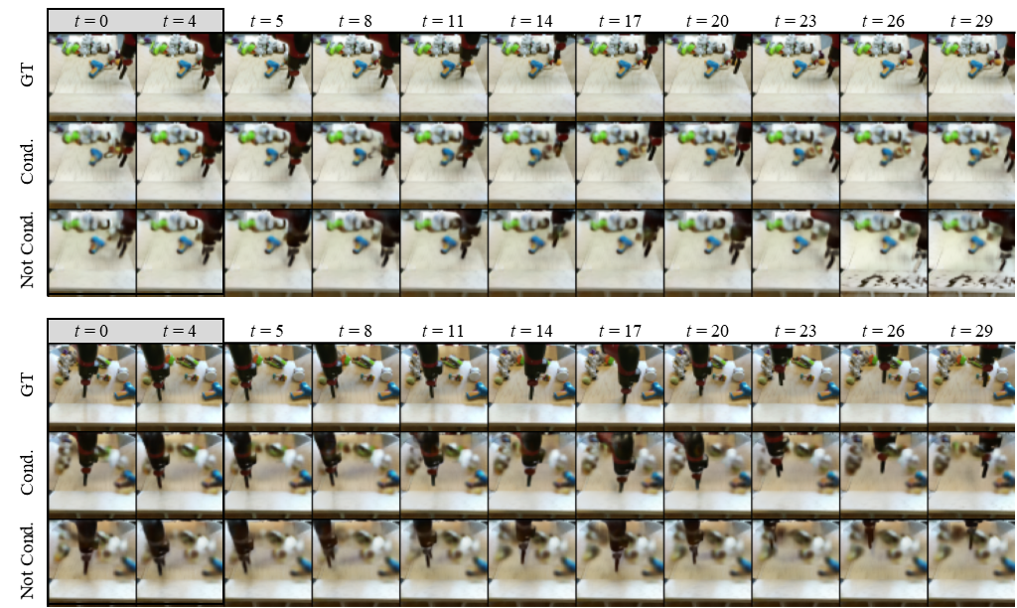}
    \vspace{0.3cm}
    \caption{Comparison between the trajectory of the agent of BAIR dataset, with and without latent dynamics sharing with keypoint module. From first to third row, groud truth observation, trajectory prediction with shared dynamics, and mask latent dynamics alone are shown.}
    \label{fig:kypt_cond_bair}
    \vspace{-0.5cm}
\end{figure}

%% file: fig/sample/04_balls_inter/01_balls_inter.tex
\begin{figure}[!ht]
    \centering
    \includegraphics[scale=1.0]{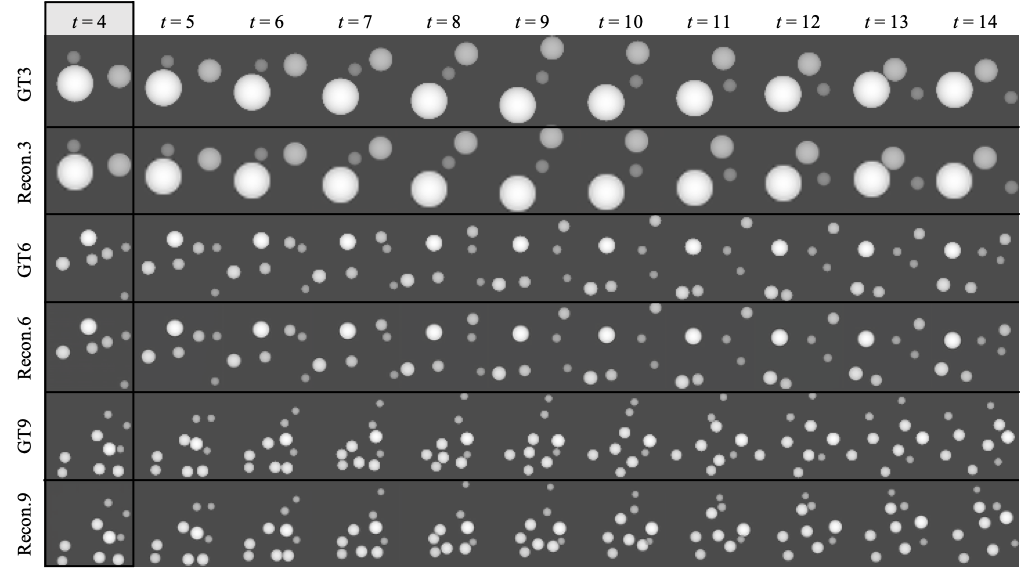}
    \caption{Samples from agent-free physical interaction among 3D balls from BALLS dataset. From top to bottom rows, the predictions using our $k$NN-based object interaction method for increasing number of balls are shown.}
        \label{fig:balls_inter}
\end{figure}